\documentclass[table]{gtech}
\PassOptionsToPackage{table, usenames, dvipsnames}{xcolor}
\usepackage{xcolor}
\usepackage{amsmath}
\usepackage{amssymb}
\usepackage[most]{tcolorbox}
\usepackage{url}
\usepackage{float}
\usepackage{enumitem}
\usepackage{subcaption}
\usepackage{caption}
\RequirePackage{tgpagella}
\RequirePackage{mathpazo}
\RequirePackage{inconsolata}
\usepackage{hyperref}
\usepackage{xurl}
\usepackage{times}
\usepackage[T1]{fontenc}
\usepackage[utf8]{inputenc}
\usepackage{CJKutf8}
\usepackage{microtype}
\usepackage{xspace}
\usepackage{booktabs}
\usepackage{graphicx}
\usepackage[export]{adjustbox}
\usepackage{colortbl}
\usepackage{listings}

\newcommand{\ignore}[1]{}

\renewcommand{\title}[1]{\newcommand{\titlelist}{{\huge\selectfont #1}}}

\title{\textbf{\raisebox{-0.18em}{\includegraphics[height=1.15em]{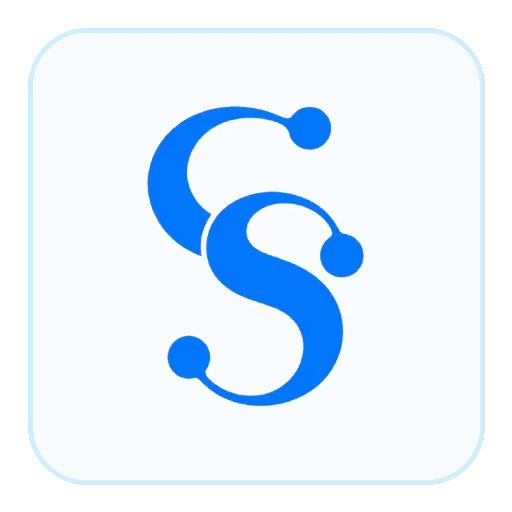}}\;SearchSwarm: Towards Delegation Intelligence in Agentic LLMs for Long-Horizon Deep Research}}

\author[1,*,\dagger]{Xiaochong Lan}
\author[2,*]{Quan Chen}
\author[3,*]{Kun Tao}
\author[4]{Xinyu Tang}
\author[3]{Tianshu Wang}
\author[3]{Qianggang Cao}
\author[3]{Xinyu Kong}
\author[3]{Zujie Wen}
\author[3]{Zhiqiang Zhang}
\author[3]{Jun Zhou}

\affiliation[1]{Tsinghua University}
\affiliation[2]{Peking University}
\affiliation[3]{Ant Group}
\affiliation[4]{Gaoling School of Artificial Intelligence, Renmin University of China}

\contribution[*]{Core contributors}
\contribution[\dagger]{Project Lead.\protect\\[2pt]Xiaochong Lan, Quan Chen, and Xinyu Tang completed this work during their internship at Ant Group.}

\renewcommand{\projectpageline}{{\normalsize\fontfamily{optimistic}\selectfont Project Page:\enspace\href{https://search-swarm.github.io}{\textcolor{black}{search-swarm.github.io}}}\\[2pt]{\normalsize\fontfamily{optimistic}\selectfont Correspondence:\enspace\href{mailto:search-swarm-2026@foxmail.com}{\textcolor{black}{search-swarm-2026@foxmail.com}}}}

\abstract{\fontsize{11pt}{12pt}\selectfont
Large language models are increasingly expected to handle complex, long-horizon real-world tasks whose context demands can grow without bound, yet model context windows remain inherently finite. Recent work explores a paradigm where a main agent decomposes tasks and dispatches subtasks to subagents, which execute and return only summarized results, conserving the main agent's context budget. However, performing this well requires \emph{delegation intelligence}: the ability to decompose complex tasks, determine when and what to delegate, and integrate returned results into the ongoing workflow. Training data for this capability is scarce in naturally occurring text, and to our knowledge, how to synthesize such data and train models to acquire this capability remains largely unexplored in the open-source community. To bridge this gap, we present a preliminary exploration targeting deep research, a representative long-horizon agent task. Specifically, we design a harness that guides the model toward high-quality task decomposition and delegation, while constraining subagents to return results properly to support the main agent's workflow. The harness-guided trajectories naturally encode correct delegation decisions, which we use as supervised fine-tuning data to internalize delegation intelligence into model weights. Our resulting model, SearchSwarm-30B-A3B, achieves 68.1 on BrowseComp and 73.3 on BrowseComp-ZH, the best results among all models of comparable scale. We will release our harness, model weights, and training data to facilitate future research.
}

\renewcommand{\secondlogocmd}{%
  \includegraphics[height=0.8cm,valign=c]{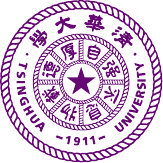}\hspace{0.3cm}%
  \includegraphics[height=0.8cm,valign=c]{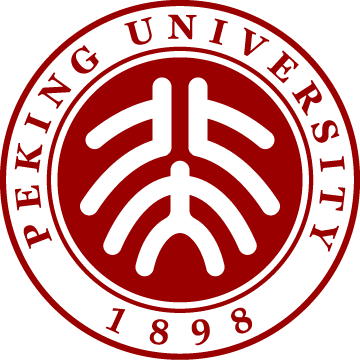}\hspace{0.3cm}%
  \includegraphics[width=3.8cm,valign=c]{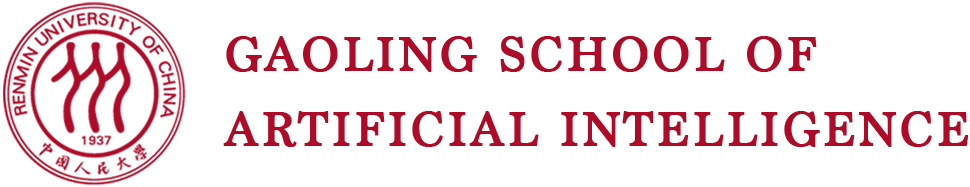}%
}
\begin{document}
\maketitle

\section{Introduction}
\label{sec:intro}

\begin{figure}[t]
    \centering
    \begin{subfigure}{0.49\linewidth}
        \centering
        \includegraphics[width=\linewidth]{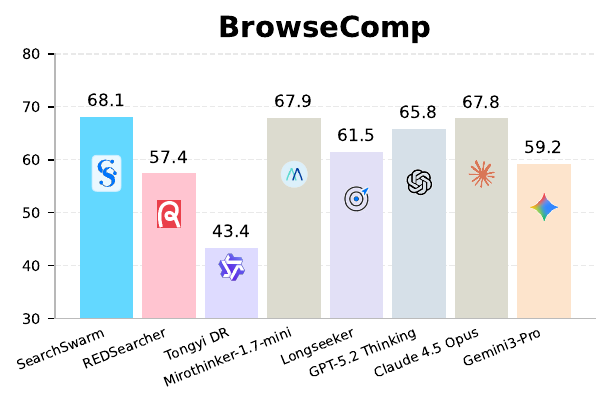}
    \end{subfigure}
    \hfill
    \begin{subfigure}{0.49\linewidth}
        \centering
        \includegraphics[width=\linewidth]{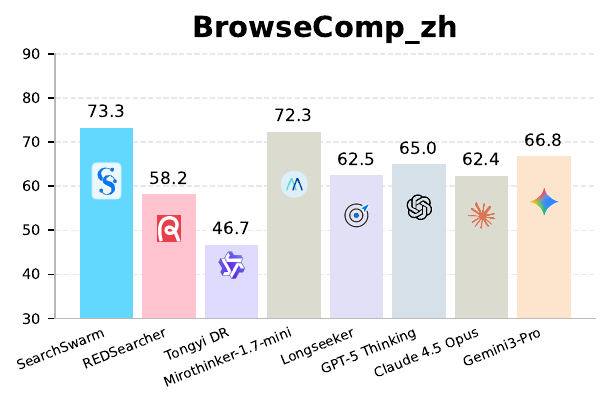}
    \end{subfigure}

    \vspace{0.6em}

    \begin{subfigure}{0.49\linewidth}
        \centering
        \includegraphics[width=\linewidth]{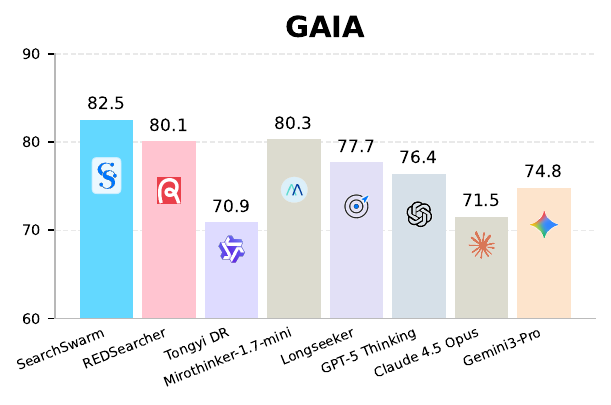}
    \end{subfigure}
    \hfill
    \begin{subfigure}{0.49\linewidth}
        \centering
        \includegraphics[width=\linewidth]{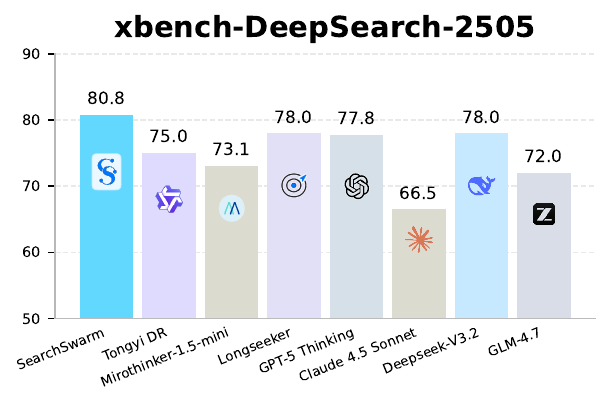}
    \end{subfigure}

    \caption{Performance comparison of SearchSwarm against lightweight models of comparable scale and larger closed-source/open-source models on four benchmarks. SearchSwarm achieves the best results among all models at the same scale and remains competitive with models over 10$\times$ larger.}
    \label{fig:four_figs}
\end{figure}

Large language models are increasingly deployed as agents for complex, long-horizon real-world tasks whose context demands can grow without bound \citep{jimenez2024swe, zhang2026repozero, yang2026programbench}, yet model context windows remain inherently finite. This fundamental tension necessitates context management strategies that selectively retain or condense information to fit within limited capacity. Early approaches include summarizing interaction history after exceeding a length threshold, or retaining only a portion of tool outputs, among others \citep{liu2025deepseek, zeng2026glm, team2026mirothinker}. However, these methods are fundamentally \emph{passive}: they lack prior planning, waiting until a context budget is exhausted before compressing, or indiscriminately discarding past observations by fixed rules.

In contrast, a paradigm where the main agent decomposes tasks and delegates subtasks to subagents represents a more \emph{active} and intelligent form of context management \citep{anthropic2025multiagent}. Rather than directly executing all steps and passively post-processing an ever-growing trajectory, the main agent plans the decomposition in advance, dispatches bounded subtasks to subagents, and receives only their summarized execution results. Several recent efforts have explored this direction with encouraging outcomes. \citet{team2026kimi} propose Agent Swarm, which freezes subagent parameters and uses reinforcement learning to train the main agent to distribute tasks effectively. \citet{huang2026step} have also reported positive results with a main-distributes, sub-executes framework. However, these efforts focus on high-level architecture and training algorithms, without providing a complete recipe covering harness design, training data construction, and model training for developing delegation intelligence.

While this paradigm is conceptually straightforward, executing it well is non-trivial. We term the requisite capability \emph{delegation intelligence}: the main agent's ability to decompose complex tasks, determine when and what to delegate to subagents, and integrate returned results into its ongoing workflow. Training data for developing delegation intelligence is scarce in naturally occurring text, as natural corpora rarely exhibit explicit multi-agent coordination. To our knowledge, how to systematically synthesize such data and train models to acquire this capability remains largely unexplored in the open-source community.

To address this, we present a preliminary exploration of constructing training data for delegation intelligence in the context of deep research, a representative long-horizon agent task. Our core idea is to first design a harness that elicits high-quality delegation behavior at inference time, and then use the resulting trajectories as a source of training data. Specifically, in our harness, the main agent dispatches work to parallel subagents via a \texttt{call\_sub\_agent} tool. Our harness encourages the main agent to delegate lower-level execution while maintaining an independent understanding of the overall research progress. Unlike similar work~\citep{huang2026wideseek, xu2026wideseek}, we require the main agent to brief each subagent with not only the task description but also the rationale, including why the subtask matters and how it fits into the broader research goal, so that the subagent can conduct focused research without redundant exploration. On the reporting side, we constrain subagent outputs to include explicit source citations, enabling the main agent to verify conclusions and propagate citations to its final response for a more trustworthy user experience.

We further leverage the harness to synthesize high-quality supervised fine-tuning data. We filter the harness-guided trajectories to retain those that encode correct delegation decisions, including when to decompose, how to scope subtasks, and how to brief subagents with appropriate context. Supervised fine-tuning internalizes these decision patterns into model weights, enabling a model that initially lacks delegation intelligence to exhibit this behavior. Our resulting model, SearchSwarm-30B-A3B, achieves 68.1 on BrowseComp and 73.3 on BrowseComp-ZH, the best results among all models of comparable scale (Figure~\ref{fig:four_figs}).

Our work makes the following contributions:
\begin{itemize}[leftmargin=*]
\item We present, to our knowledge, one of the first explorations of \emph{delegation intelligence} for super-long-horizon agent tasks, and design a harness for the main-distributes, sub-executes paradigm that guides the main agent to intelligently decompose a task and dispatch subtasks to independent-context subagents, improving deep research performance.
\item Using the harness, we synthesize high-quality supervised fine-tuning data that internalizes delegation behavior into model weights. The resulting model, SearchSwarm, achieves 68.1 on BrowseComp, 73.3 on BrowseComp-ZH, 82.5 on GAIA, and 80.8 on xbench-DeepSearch, the best results among all models of comparable scale.
\item We open-source our harness, model weights, and training data, to facilitate future research on delegation intelligence and multi-agent coordination.
\end{itemize}
\section{Method}
\label{sec:method}

\begin{figure}[t]
    \centering
    \includegraphics[width=\linewidth]{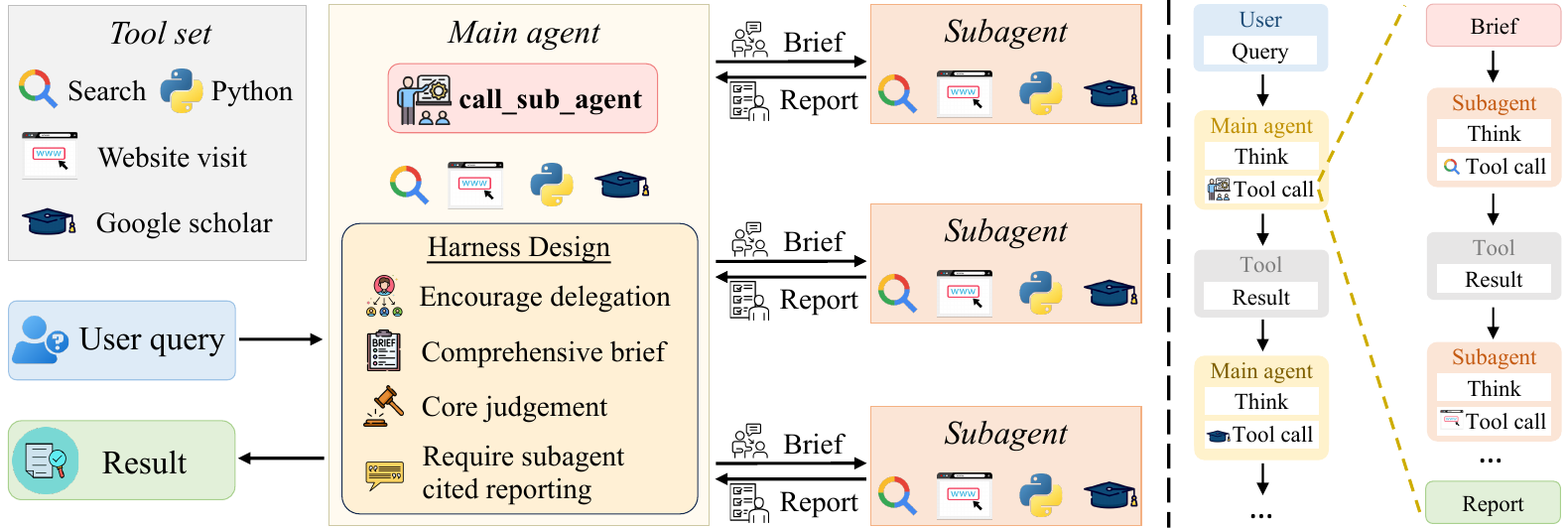}
    \caption{Overview of SearchSwarm. \textbf{Left:} System architecture. The main agent receives a user query and is equipped with standard information retrieval tools and the \texttt{call\_sub\_agent} delegation tool. The harness design guides the main agent through four principles: encouraging delegation, writing comprehensive briefs, retaining core judgment, and requiring citation-grounded subagent reporting. Each subagent operates in an independent context with standard tools, receiving only a brief and returning a report. \textbf{Right:} Execution flow of a research session. The main agent alternates between reasoning and tool calls; upon invoking \texttt{call\_sub\_agent}, a subagent executes the subtask through its own multi-turn tool interactions in a separate context, and returns a condensed report that re-enters the main agent's context for further reasoning.}
    \label{fig:overview}
\end{figure}

SearchSwarm follows a main-distributes, sub-executes paradigm, illustrated in Figure~\ref{fig:overview}: the main agent plans and delegates bounded subtasks to independent subagents, then integrates their condensed reports. We first formalize the setting (Section~\ref{sec:formulation}), then describe the harness that elicits high-quality delegation (Section~\ref{sec:harness}), and finally how its trajectories are internalized into model weights via supervised fine-tuning.

\subsection{Formulation}
\label{sec:formulation}

We model the deep research task as a multi-turn interaction between an agent and a tool-equipped environment. Given a user question $q$, the agent issues tool calls over multiple steps to gather information and produces an evidence-grounded answer $y$. We adopt the ReAct \citep{yao2022react} framework to organize the interaction. Each step $t$ consists of three components:
\begin{itemize}[leftmargin=*]
\item \textbf{Thought} ($\tau_t$): The agent's internal reasoning, including analyzing available evidence, identifying information gaps, assessing the plausibility of current hypotheses, and planning the next action. $\tau_t$ serves as a compact representation of the interaction history that guides action selection.
\item \textbf{Action} ($a_t$): A tool call executed by the agent. The action space includes standard information retrieval tools and \texttt{call\_sub\_agent}.
\item \textbf{Observation} ($o_t$): The result returned by the environment after executing $a_t$.
\end{itemize}

A complete trajectory is recorded as:
\begin{equation}
H_T = \bigl(q,\; (\tau_0, a_0, o_0),\; \ldots,\; (\tau_T, a_T, o_T),\; y\bigr).
\end{equation}
At each step, thought and action are sampled from the policy:
\begin{equation}
\tau_t, a_t \sim \pi(\cdot \mid q, H_{t-1}).
\end{equation}
The final answer is generated from the accumulated evidence: $y = g(q, H_T)$. When evidence is incomplete or contradictory, $y$ should explicitly reflect uncertainty.

\paragraph{Delegation.}
When $a_t = \texttt{call\_sub\_agent}(b)$, the agent delegates a subtask for execution. The brief $b$ contains a subtask description and relevant context extracted from the agent's current reasoning. It triggers an independent sub-trajectory:
\begin{equation}
H^{\text{sub}} = \bigl(b,\; (\tau_0^s, a_0^s, o_0^s),\; \ldots,\; (\tau_S^s, a_S^s, o_S^s),\; r\bigr),
\end{equation}
which executes in a separate context conditioned solely on $b$, with no visibility into the main agent's history $H_{t-1}$. Upon completion, the sub-trajectory produces a report $r$, and the main agent receives:
\begin{equation}
o_t = r.
\end{equation}
The main agent observes only the final report; the intermediate steps of $H^{\text{sub}}$ are not visible.

\paragraph{Delegation as context management.}
In long-horizon tasks, the agent's context grows continuously as tool calls accumulate, necessitating management strategies. Existing approaches address this through various mechanisms: discarding history beyond a threshold, retaining only the most recent few rounds of tool calls, or compressing the trajectory into a summary~\citep{liu2025deepseek, zeng2026glm, team2026mirothinker}.

Although our method dispatches work to sub-agents, it involves only a single model: the sub-agents are the same model invoked in independent, fresh contexts, not separate or additional models. When \texttt{call\_sub\_agent} is invoked, the next reasoning step is conditioned only on the brief $b$, not the full history $H_{t-1}$, retaining only the information the agent deems essential for the subtask; after execution completes, what re-enters the main context is the report $r$, a compressed summary of the entire sub-trajectory. Both the brief and the report are generated by the model, rather than determined by fixed rules. Our approach can thus be considered as single-agent context management rather than a multi-agent system: the only difference from prior context-management methods is that the model manages its own context more intelligently, using the model-generated brief and report as a content-aware compression in place of fixed-rule truncation or summarization. Comparisons with such methods are therefore made on equal footing.

\subsection{Harness Design}
\label{sec:harness}

We design a harness comprising a tool set and system prompts for the main agent and subagents that guides an LLM toward high-quality delegation behavior. This section describes the tool interface and core design principles. Full system prompts are provided in Appendix~\ref{sec:appendix-prompts}.

\paragraph{Tools.}
The agent is equipped with the following tools: \texttt{search} submits queries to a search engine and returns ranked results with titles, URLs, and snippets; \texttt{visit} accesses a specified URL and extracts page content; \texttt{google\_scholar} retrieves academic literature; \texttt{python} provides a code execution environment for numerical computation and data processing. These form the base information retrieval capabilities. On top of them, we introduce \texttt{call\_sub\_agent} as the core delegation tool: the main agent submits a brief, and the subagent executes in an independent context and returns a report. Subagents are equipped with the same standard tools but do not have access to \texttt{call\_sub\_agent}, limiting delegation to a single level.

\paragraph{Encouraging delegation.}
Because the main agent's context is finite, every token it spends on raw \texttt{search} and \texttt{visit} outputs competes directly with the planning, verification, and synthesis that only it can perform. Multi-step information gathering is precisely the kind of work that is token-expensive but cognitively shallow: it can take many turns to surface a single fact. The harness therefore directs the main agent to hand such gathering to subagents, which pay the exploration cost in their own contexts and return only a condensed result, keeping the main agent's limited attention on high-level coordination. The main agent gathers information itself only when a subtask is shallow enough that the overhead of delegating would outweigh the context it saves.

\paragraph{Comprehensive briefing.}
Subagents start in a fresh context with no knowledge of prior investigation progress. The brief is the sole channel through which a subagent receives context, and its quality directly determines subagent effectiveness. When a brief contains only a simple task instruction, subagents tend to search aimlessly or re-investigate facts the main agent has already confirmed, producing results that fail to advance the overall investigation. We therefore require the main agent to write each brief as if addressing a new collaborator joining the investigation: beyond the subtask description, the brief includes why this subtask matters to the overall question, what has been established so far, what remains uncertain, and which directions have been tried or ruled out. This aligns the subagent with the main agent's research progress, ensuring its output contributes maximally to the overall investigation.

\paragraph{Main agent retains core judgment.}
The main agent is the only entity with a complete view across all subtasks, and only it can judge whether a subagent's findings are consistent with other findings and the overall evidence landscape. If subagent reports are trusted without scrutiny, errors propagate and accumulate, undermining the coherence of the overall reasoning. The harness therefore requires subagents to focus on gathering evidence and testing specific hypotheses, while all directional decisions are made independently by the main agent, including which hypothesis to pursue, when to terminate the investigation, and how to adjudicate between conflicting reports.

\paragraph{Citation-grounded reporting.}
Under the delegation architecture, the main agent cannot observe a subagent's intermediate execution. If a subagent's report does not cite its sources, the main agent cannot distinguish well-supported conclusions from hallucinations or misinterpretations. We therefore require subagent reports to attach inline citations to every important conclusion, pointing to specific source URLs, enabling the main agent to verify the reliability of reported findings. The main agent's final response likewise includes inline citations, providing end-to-end traceability from sources to conclusions for the user.

\subsection{Supervised Fine-tuning}

\paragraph{Data Collection.}
To train a model that can both delegate effectively and execute delegated tasks, we require trajectories exhibiting both behaviors. We source queries from the open-source RedSearcher \citep{chu2026redsearcher} and OpenSeeker \citep{du2026openseeker} datasets. The model executes deep research tasks on these queries under harness guidance, and we record the complete execution trajectories, including thinking, tool calls, and environment returns, as training data. We use two configurations for trajectory collection. In the first, a single model serves as both main agent and subagent, and trajectories from both roles are retained. In the second, a stronger model serves as the main agent paired with a weaker subagent, and only main agent trajectories are retained. The rationale for the second configuration is that less reliable subagent results force the main agent to exercise tighter control over the research mainline, producing trajectories with more deliberate task decomposition and more rigorous result verification. Data from both configurations are mixed to form the training set. The main agent context window is set to 128K tokens and the subagent to 64K. When a trajectory approaches the context limit, the model is prompted to produce a final answer immediately. We retain these forced-answer trajectories rather than discarding them, so that the model learns to deliver high-quality responses when the same forced-answer mechanism is triggered at test time.

\paragraph{Filtering.}
We retain only main agent trajectories with correct final answers. Subagent trajectories are kept only when the corresponding main trajectory is correct, and overly short subagent trajectories are downsampled. Samples containing undesirable behavior patterns are removed, including repeated identical tool calls, hallucinated citations to nonexistent sources, and tool misuse such as web access attempts through the \texttt{python} interpreter.

\paragraph{Training Objective.}
Let a trajectory $\tau = (a_1, o_1, a_2, o_2, \ldots, a_T, o_T)$ consist of alternating model outputs $a_t$ (thinking and tool calls) and environment returns $o_t$ (tool results, including subagent reports). We fine-tune the base model via next-token prediction with environment masking:
\begin{equation}
\mathcal{L} = -\sum_{t=1}^{T} \sum_{j=1}^{|a_t|} \log p_\theta\bigl(a_t^{(j)} \mid a_t^{(<j)}, \tau_{<t}\bigr)
\end{equation}
where $a_t^{(j)}$ is the $j$-th token of the model output at step $t$, and $\tau_{<t} = (a_1, o_1, \ldots, a_{t-1}, o_{t-1})$ is the preceding context. The loss is computed only over model outputs $a_t$; all environment returns $o_t$ are masked. This applies uniformly to both main agent and subagent trajectories, training the model to produce appropriate reasoning and tool invocations given the observed context without memorizing environment content.
\section{Experiments}
\label{sec:experiments}

\subsection{Experimental Setup}

\paragraph{Benchmarks.}
We evaluate on four challenging benchmarks representative of long-horizon research tasks: BrowseComp \citep{wei2025browsecomp}, BrowseComp-ZH \citep{zhou2025browsecomp}, GAIA \citep{mialon2024gaia}, and xbench-DeepSearch-2505 \citep{xbench2025}. We follow the evaluation method of \citet{team2026mirothinker}. We use DeepSeek-V4-Flash as the judge model and manually verify the correctness of its judgments. For BrowseComp-ZH, we use the corrected version provided by \citet{team2026longcat}.

\paragraph{Baselines.}
We compare against three categories of models. \textit{Closed-source models}: GPT-5.2-Thinking \citep{openai2025gpt52}, GPT-5 \citep{openai2025gpt5}, Claude-4.5-Sonnet \citep{anthropic2025sonnet45}, Claude-4.5-Opus \citep{anthropic2025opus45}, Gemini-3.0-Pro \citep{google2025gemini3}, and Seed-2.0-Pro \citep{bytedance2026seed2}. \textit{Open-source models}: DeepSeek V3.2 \citep{liu2025deepseek}, GLM-4.7 \citep{glm2025glm47}, GLM-5.0 \citep{zeng2026glm}, MiniMax-M2 \citep{minimax2025m2}, MiniMax-M2.5 \citep{minimax2026m25}, Kimi-K2.5 \citep{team2026kimi}, LongCat-Flash-Thinking-2601 \citep{team2026longcat}, and Step-3.5-Flash \citep{huang2026step}. \textit{Open-source lightweight models} at the same 30B-A3B scale: Tongyi DeepResearch \citep{team2025tongyi}, RedSearcher \citep{chu2026redsearcher}, LongSeeker \citep{lu2026longseeker}, MiroThinker-1.5-mini, and MiroThinker-1.7-mini \citep{team2026mirothinker}.

\paragraph{Implementation Details.}
We fine-tune the base model, Tongyi DeepResearch-30B-A3B~\citep{team2025tongyi}, with a batch size of 128. The learning rate decays from 5e-5 to 1e-6 following a cosine schedule. During inference, we set the temperature to 0.85, top\_p to 0.95, and presence penalty to 1.1. The maximum context length is 128K tokens for the main agent and 64K tokens for the subagent. The maximum generation length per turn is 8K tokens for both roles. When either agent's context exceeds its limit, we roll back to the previous round and force the model to produce a final answer. We explicitly inform the agent that the \texttt{python} interpreter is stateless: variables and imports from previous calls are not preserved across turns. For the \texttt{search} tool, we use the Serper API, returning 10 results per query. For the \texttt{visit} tool, we use Jina for web page content extraction.

\subsection{Main Results}

\begin{table}[t]
\centering
\caption{Main results. Baseline results are collected from respective technical reports or model cards. * indicates results with context management. \textbf{Bold} indicates the best result among open-source lightweight models.}
\label{tab:main-results}
\adjustbox{max width=\linewidth}{
\begin{tabular}{lccccc}
\toprule
\textbf{Model} & \textbf{Size} & \textbf{BrowseComp} & \textbf{BrowseComp-ZH} & \textbf{GAIA} & \textbf{xbench-DeepSearch-2505} \\
\midrule
\rowcolor{gray!8} \multicolumn{6}{l}{\textit{Closed-source models}} \\
GPT-5.2-Thinking & -- & 65.8 & 76.1 & -- & -- \\
GPT-5 & -- & 54.9 & 65.0 & 76.4 & 77.8 \\
Claude-4.5-Opus & -- & 67.8 & 62.4 & 71.5 & -- \\
Claude-4.5-Sonnet & -- & 24.1 & 42.4 & 66.0 & 66.5 \\
Gemini-3.0-Pro & -- & 59.2 & 66.8 & 74.8 & -- \\
Seed-2.0-Pro & -- & 77.3* & 82.4* & 78.6 & -- \\
\midrule
\rowcolor{gray!8} \multicolumn{6}{l}{\textit{Open-source models}} \\
Kimi-K2.5 & 1T-A32B & 78.4* & -- & -- & -- \\
GLM-4.7 & 355B-A32B & 67.5* & 66.6* & -- & 72.0 \\
GLM-5.0 & 744B-A40B & 75.9* & 72.7* & -- & -- \\
DeepSeek V3.2 & 671B-A37B & 67.6* & 65.0* & 75.1 & 78.0 \\
LongCat-Flash-Thinking-2601 & 560B-A27B & 73.1* & 77.7* & -- & -- \\
MiniMax-M2 & 230B-A10B & 44.0 & -- & 75.7 & 72.0 \\
MiniMax-M2.5 & 230B-A10B & 76.3* & -- & -- & -- \\
Step-3.5-Flash & 196B-A11B & 69.0* & 66.9 & 84.5 & 83.7 \\
\midrule
\rowcolor{gray!8} \multicolumn{6}{l}{\textit{Open-source lightweight models}} \\
Tongyi DeepResearch & 30B-A3B & 43.4 & 46.7 & 70.9 & 75.0 \\
Tongyi DR Swarm & 30B-A3B & $\approx$43.4 & $\approx$46.7 & $\approx$70.9 & $\approx$75.0 \\
RedSearcher & 30B-A3B & 57.4* & 58.2* & 80.1 & -- \\
LongSeeker & 30B-A3B & 61.5* & 62.5* & 77.7* & 78.0* \\
MiroThinker-1.5-mini & 30B-A3B & 56.1* & 66.8* & 72.0* & 73.1* \\
MiroThinker-1.7-mini & 30B-A3B & 67.9* & 72.3* & 80.3* & -- \\
\midrule
\rowcolor{blue!5} \textbf{SearchSwarm (Ours)} & 30B-A3B & \textbf{68.1*} & \textbf{73.3*} & \textbf{82.5*} & \textbf{80.8*} \\
\bottomrule
\end{tabular}
}
\end{table}

Table~\ref{tab:main-results} presents the main results. As established in Section~\ref{sec:formulation}, our delegation mechanism can be understood as a form of context management from the main agent's perspective: each subagent call resets the working context to a brief, and only a compressed report re-enters the main context upon completion. This places our method in the same category as other context management approaches, making comparisons with models employing such techniques (marked with *) fair.

SearchSwarm achieves state-of-the-art performance among all 30B-A3B scale models across all four benchmarks. It surpasses MiroThinker-1.7-mini, the previous best at this scale, on BrowseComp (68.1 vs.\ 67.9), BrowseComp-ZH (73.3 vs.\ 72.3), and GAIA (82.5 vs.\ 80.3). On xbench-DeepSearch-2505, SearchSwarm (80.8) exceeds LongSeeker (78.0) and Tongyi DeepResearch (75.0). Compared to the base model without context management (43.4 on BrowseComp), our training yields a 24.7-point absolute improvement, demonstrating the substantial impact of delegation intelligence.

Beyond the lightweight category, SearchSwarm exhibits strong competitiveness against models of substantially larger scale. On BrowseComp, it matches DeepSeek V3.2 (671B-A37B, 67.6) and exceeds GPT-5.2-Thinking (65.8). On GAIA, SearchSwarm (82.5) surpasses GPT-5 (76.4) and Seed-2.0-Pro (78.6), falling short only of Step-3.5-Flash (84.5). These results suggest that well-trained delegation intelligence enables a lightweight model to achieve performance competitive with frontier systems on long-horizon research tasks.

We additionally report Tongyi DR Swarm, which applies our harness to the base Tongyi DeepResearch model without fine-tuning. We observe that this model \textit{never} invokes the \texttt{call\_sub\_agent} tool, behaving identically to Tongyi DeepResearch without the harness. We therefore report the results from the official Tongyi DeepResearch technical report. This confirms that delegation behavior does not emerge from the harness alone and requires explicit training.

\subsection{Effectiveness of the Harness}

We conduct an ablation study to validate the effectiveness of our harness design. On a 200-question subset of BrowseComp, we compare DeepSeek V3.2 under three configurations: (1) the original Tongyi DeepResearch framework, which scores 47.7; (2) the Tongyi DeepResearch framework augmented with the \texttt{call\_sub\_agent} tool with only its parameter schema described, which scores 50.0; and (3) our full harness, which scores 57.7. Simply providing the delegation tool yields a modest improvement (+2.3), but the full harness with its design principles for encouraging delegation, comprehensive briefing, and citation-grounded reporting produces a substantially larger gain (+10.0 over the base framework). Analysis of model behavior confirms that subagent invocation frequency increases significantly under the full harness. These results demonstrate the effectiveness of our harness in eliciting intelligent delegation behavior.

\subsection{Training from a Different Base Model}

We adopt Tongyi DeepResearch as the base model for our main experiments primarily to emphasize our contribution in constructing training data for delegation intelligence, and to obtain a versatile model that can both delegate to subagents and complete tasks entirely on its own. To isolate the contribution of the training data itself, we additionally fine-tune Qwen3-30B-A3B-Thinking-2507 on exactly the same data. Under the same experimental setup as our main experiments, this model attains 66.5 on a 200-question subset of BrowseComp and 64.0 on BrowseComp-ZH, exceeding the results that RedSearcher and LongSeeker report under their respective best context-management configurations.\footnote{LongSeeker also reports on a 200-question subset, so its result is directly comparable to ours. RedSearcher reports on the complete 1266-question BrowseComp benchmark; although a 200-question subset is not identical, our margin is large enough that our model is very likely the stronger one. Our main results in Table~\ref{tab:main-results} use the complete BrowseComp benchmark.} Since Qwen3-30B-A3B-Thinking-2507 has not been optimized for deep search at all, this result shows that our training data alone can train a strong deep-research model, attesting to the quality of both the training data and the harness on which its synthesis and evaluation are built.

\subsection{Generalization to the Single-Agent Setting}

We evaluate whether the capabilities acquired through delegation training generalize to a setting where the \texttt{call\_sub\_agent} tool is not available. We compare our model and Tongyi DeepResearch under an identical single-agent configuration: a single 128K-token context, no context management, and the \texttt{call\_sub\_agent} tool disabled. On the same 200-question BrowseComp subset and BrowseComp-ZH, our model achieves 52.0 and 53.3, improving over Tongyi DeepResearch (43.5 and 46.5) on both. Notably, our training data does not include any trajectories collected without the subagent tool. The improvement suggests that the intelligence embodied in our training data, including systematic problem decomposition, methodical resolution of sub-questions, and maintenance of overall research progress, generalizes beyond the delegation setting and benefits the model even when it must execute all steps itself.

\subsection{Generalization to Open-Ended Deep Research}

\begin{table}[t]
\centering
\caption{Results on open-ended deep research benchmarks. Due to resource constraints, we evaluate on a 200-question subset of HealthBench and ResearchQA. Average is computed only when all four scores are available. \textbf{Bold} indicates the best result among open-source models.}
\label{tab:open-ended}
\adjustbox{max width=\linewidth}{
\begin{tabular}{lccccc}
\toprule
\textbf{Model} & \textbf{ScholarQA-v2} & \textbf{HealthBench} & \textbf{ResearchQA} & \textbf{DeepResearchBench} & \textbf{Average} \\
\midrule
\rowcolor{gray!8} \multicolumn{6}{l}{\textit{Closed-source systems}} \\
OpenAI DeepResearch & 79.6 & 53.8 & 79.2 & 46.9 & 64.9 \\
Perplexity DeepResearch & 67.3 & -- & 75.3 & 42.3 & -- \\
Gemini-3.1-Pro + search & -- & 47.5 & 74.5 & 44.4 & -- \\
\midrule
\rowcolor{gray!8} \multicolumn{6}{l}{\textit{Open-source models}} \\
Qwen3-8B & 40.4 & 16.5 & 56.1 & 33.3 & 36.6 \\
QwQ-32B & 41.9 & 24.5 & 60.9 & 40.3 & 41.9 \\
Tongyi DeepResearch & 46.5 & 46.2 & 66.7 & 40.6 & 50.0 \\
WebThinker-32B-DPO & 46.7 & 39.4 & 74.2 & 40.6 & 50.2 \\
Dr.Tulu & \textbf{88.3} & \textbf{52.8} & 75.7 & \textbf{45.4} & \textbf{65.6} \\
\midrule
\rowcolor{blue!5} \textbf{SearchSwarm (Ours)} & 79.2 & \textbf{52.8} & \textbf{80.2} & 44.4 & 64.2 \\
\bottomrule
\end{tabular}
}
\end{table}

The benchmarks in Section~\ref{sec:experiments} focus on short-answer information retrieval tasks. To assess whether our model generalizes to tasks requiring long-form, synthesized responses, we additionally evaluate on four open-ended deep research benchmarks: ScholarQA-v2, HealthBench, ResearchQA, and DeepResearchBench. We follow the evaluation protocol of \citet{shao2025dr}. Due to resource constraints, we evaluate on a 200-question subset of HealthBench and ResearchQA. Table~\ref{tab:open-ended} presents the results.

SearchSwarm substantially outperforms its base model, Tongyi DeepResearch, across all four benchmarks, with an average improvement of 14.2 points (64.2 vs.\ 50.0). The gains are particularly evident on ScholarQA-v2 (+32.7) and ResearchQA (+13.5), both of which require comprehensive multi-source synthesis. Among open-source models, SearchSwarm achieves the second-highest average, closely trailing Dr.Tulu (65.6) while outperforming WebThinker-32B-DPO (50.2) by a wide margin. Compared to closed-source systems, SearchSwarm approaches OpenAI DeepResearch (64.9) and exceeds Perplexity DeepResearch.

Notably, our training data contains exclusively short-answer deep research queries; no open-ended tasks are included. The generalization to open-ended settings likely stems from two aspects of our training regime. First, the delegation training teaches the model to decompose complex questions into focused subtasks and explore multiple hypotheses in parallel through subagents. This structured investigative process transfers naturally to open-ended research, where thoroughness and breadth of coverage are essential. Second, even on short-answer tasks, our harness requires the main agent to produce a comprehensive explanation with inline citations, and subagents to deliver detailed reports grounding every conclusion in retrieved evidence. This emphasis on completeness and evidence-grounded reasoning during training cultivates the model's ability to generate well-organized, long-form responses that open-ended tasks demand.

\subsection{Model Behavior Analysis}

\begin{figure}[t]
\centering
\begin{subfigure}{0.7\linewidth}
    \centering
    \includegraphics[width=\linewidth]{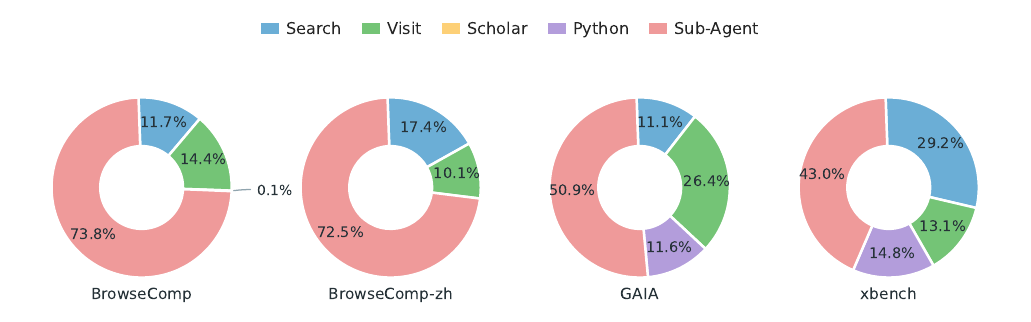}
    \caption{Main agent}
    \label{fig:tool-usage-main}
\end{subfigure}

\vspace{0.9em}

\begin{subfigure}{0.7\linewidth}
    \centering
    \includegraphics[width=\linewidth]{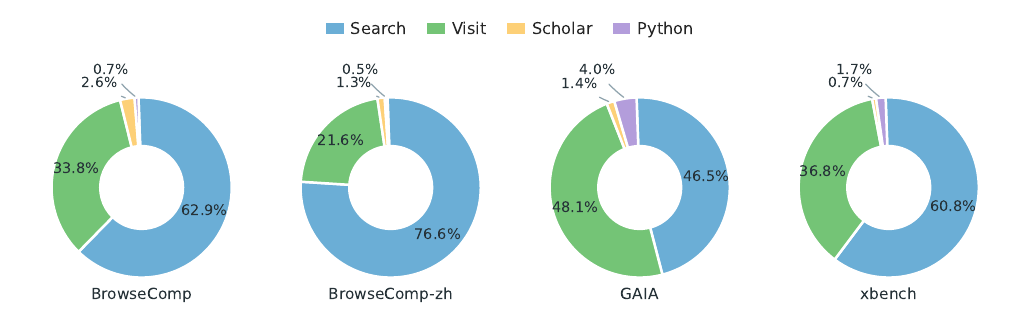}
    \caption{Subagent}
    \label{fig:tool-usage-sub}
\end{subfigure}
\caption{Tool usage distribution on four benchmarks. (a)~The main agent delegates extensively via \texttt{call\_sub\_agent}; its direct tool use is dominated by visit for verification. (b)~Subagents focus on search and visit for information gathering.}
\label{fig:tool-usage}
\end{figure}

To understand whether the model has internalized the intended delegation patterns, we analyze the tool usage distribution of the main agent and subagents across all four short-answer benchmarks. Figure~\ref{fig:tool-usage} presents the results.

\paragraph{The main agent operates primarily as an orchestrator.} \texttt{call\_sub\_agent} is the most frequently invoked tool by the main agent across all datasets, accounting for over 70\% on BrowseComp and BrowseComp-ZH, and 43--51\% on GAIA and xbench, confirming that the model has learned to delegate information gathering rather than executing searches itself.

\paragraph{Direct tool use by the main agent is verification-oriented.} When the main agent invokes tools directly, \texttt{visit} is disproportionately prominent relative to \texttt{search}---on GAIA, \texttt{visit} calls (26.4\%) substantially exceed \texttt{search} calls (11.1\%). This arises because the main agent tends to follow citation URLs from subagent reports to verify conclusions rather than initiating new searches. Subagents exhibit the opposite pattern: \texttt{search} consistently dominates (46.5--76.6\%), reflecting their role in exploratory retrieval.

\paragraph{Tool distributions reflect task characteristics.} GAIA and xbench require mathematical computation, reflected in higher \texttt{python} usage by the main agent (11.6\% and 14.8\%) and subagents (4.0\% and 1.7\%), while BrowseComp tasks show negligible \texttt{python} usage. The main agent's \texttt{python} usage is consistently higher than the subagents', suggesting the model handles computation directly while delegating search-intensive work.

Additional behavioral statistics, broken down by whether each question is ultimately answered correctly, are provided in Appendix~\ref{sec:appendix-behavior}.
\section{Related Work}
\label{sec:related}

\subsection{Delegation Intelligence}

Delegation is a fundamental strategy by which human intelligence manages complexity \citep{simon2013administrative}. Individual cognitive resources are limited, yet real-world tasks can far exceed any individual's processing capacity \citep{kahneman1973attention}. By entrusting subtasks to others and integrating their results, humans collaboratively accomplish work well beyond individual capability \citep{march1993organizations}. Effective delegation is not mere task forwarding: the delegator must judge when to delegate, how to scope subtasks, what context to provide so the delegatee can work independently, and how to integrate returned results into the overall workflow \citep{castelfranchi1998towards}.

For LLM agents, the context window constitutes an analogous resource constraint. When a task's information demands exceed the capacity of a single context, the model similarly benefits from delegation: offloading subtasks to independent agent instances and receiving only condensed results. Recent work has begun exploring this direction. \citet{anthropic2025multiagent} describe a multi-agent architecture in which a coordinator dispatches focused subagents in parallel and synthesizes their reports. \citet{team2026kimi} introduce Agent Swarm, training the main agent's task allocation policy via reinforcement learning while keeping subagent weights frozen. \citet{huang2026step} similarly adopt a hierarchical design in which a main agent dispatches subtasks to subagents for execution. \citet{ruan2026aorchestra} propose a unified four-tuple agent abstraction that enables dynamic sub-agent creation for diverse agentic tasks, providing detailed harness design and exploring supervised fine-tuning for orchestration. However, most of these efforts focus on high-level architecture and training algorithms, without providing a complete recipe covering harness design, training data construction, and model training for developing delegation intelligence. Our work presents a preliminary exploration in this direction, achieving state-of-the-art results among models of the same scale while openly releasing the full recipe, including the harness, training data, and model weights.

\subsection{Agentic Large Language Models}
Large language models are evolving from chat models to agents. Beyond single-turn question answering, models are now expected to use tools, interact with their environment, adjust strategies based on feedback, and complete complex tasks through multi-turn observation-action loops. The community has trained capable models, including Claude 4.7 \citep{anthropic2026claude47}, GPT 5.5 \citep{openai2026gpt55}, Gemini 3.1 \citep{google2026gemini31}, DeepSeek V4 \citep{deepseek2026v4}, Qwen 3.7 \citep{qwen2026qwen37}, GLM 5.1 \citep{glm2026glm51}, Kimi 2.6 \citep{kimi2026k26}, and Ring 2.6 \citep{inclusionai2026ring26}, achieving strong performance on benchmarks for coding, search, and general tool use. As tasks grow more complex with potentially unbounded context demands, delegating subtasks to subagents offers a principled way to manage the main agent's context budget. Our work is among the earliest open-source contributions in this direction.

\subsection{Search Agents}
The parameter capacity of a language model is smaller than the totality of world knowledge expressible in language, making it inherently a lossy compressor \citep{deletang2024language, allen2024physics}. Moreover, model parameters are fixed after training and cannot capture real-time information. The value of information lies in improving decisions \citep{howard1966information}, and for many decisions, long-tail and real-time information is critical. Search serves as the model's primary interface for accessing such information. For a specific retrieval task, constructing effective queries requires the model to have some familiarity with the knowledge neighborhood of the answer, yet the model's world knowledge is incomplete \citep{belkin1980anomalous}. Consequently, multi-round iterative search is often necessary, where the model progressively approaches the target information through incremental refinement. This motivates the development of search agents capable of multi-turn iterative retrieval. Existing work, including Tongyi DeepResearch \citep{team2025tongyi}, RedSearcher \citep{chu2026redsearcher}, MiroThinker \citep{team2026mirothinker}, and OpenSeeker \citep{du2026openseeker}, has explored data construction, tool design, and training pipelines for building search agents. Building on this line of work, our approach equips the main agent with subagents as callable tools, each independently handling a coherent subtask and returning only a completion summary. This shields the main agent's context from raw tool outputs, freeing capacity for broader exploration.
\section{Conclusion}
\label{sec:conclusion}

We present SearchSwarm, a preliminary exploration of training delegation intelligence for long-horizon agent tasks, demonstrated effective on deep research. We design a harness that guides the main agent toward effective task decomposition, comprehensive subagent briefing, and citation-grounded result integration, and demonstrate that this harness improves deep research performance at inference time. Using the harness, we synthesize supervised fine-tuning data that internalizes delegation behavior into model weights. The resulting model, SearchSwarm-30B-A3B, achieves state-of-the-art performance among models of comparable scale on BrowseComp, BrowseComp-ZH, GAIA, and xbench-DeepSearch, while remaining competitive with models over 10$\times$ larger. Analysis shows that the delegation intelligence acquired through training generalizes to single-agent settings and open-ended research tasks, suggesting that the structured investigative patterns encoded in our training data confer benefits beyond the specific delegation paradigm. We hope that our released harness, model weights, and training data will facilitate future research on delegation intelligence and multi-agent coordination for complex agent tasks.

\section*{Acknowledgments}
We thank Zhixun Li, Leqi Zheng, Jinbo Su, and Xiufeng Huang for their helpful discussions.

\bibliographystyle{assets/plainnat}
\bibliography{main}

\clearpage
\appendix
\section{Behavioral Analysis}
\label{sec:appendix-behavior}

We present additional behavioral statistics separated by whether the question is ultimately answered correctly (blue) or incorrectly (red).

\begin{figure*}[p]
\centering
\begin{subfigure}{0.42\textwidth}
    \centering
    \includegraphics[width=\textwidth]{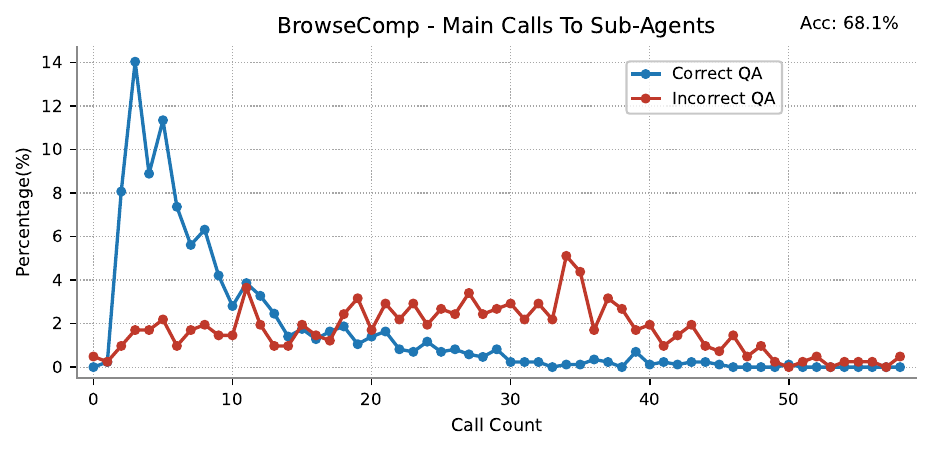}
\end{subfigure}
\hspace{0.5em}
\begin{subfigure}{0.42\textwidth}
    \centering
    \includegraphics[width=\textwidth]{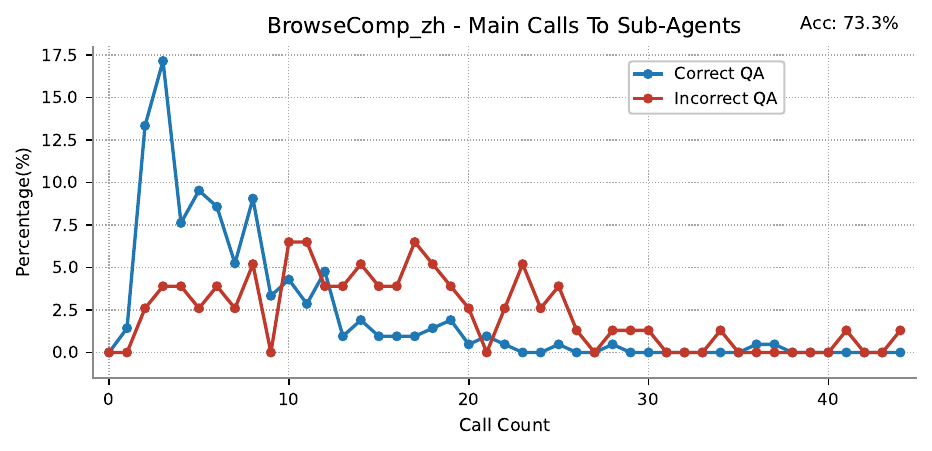}
\end{subfigure}

\vspace{-0.5em}
\begin{subfigure}{0.42\textwidth}
    \centering
    \includegraphics[width=\textwidth]{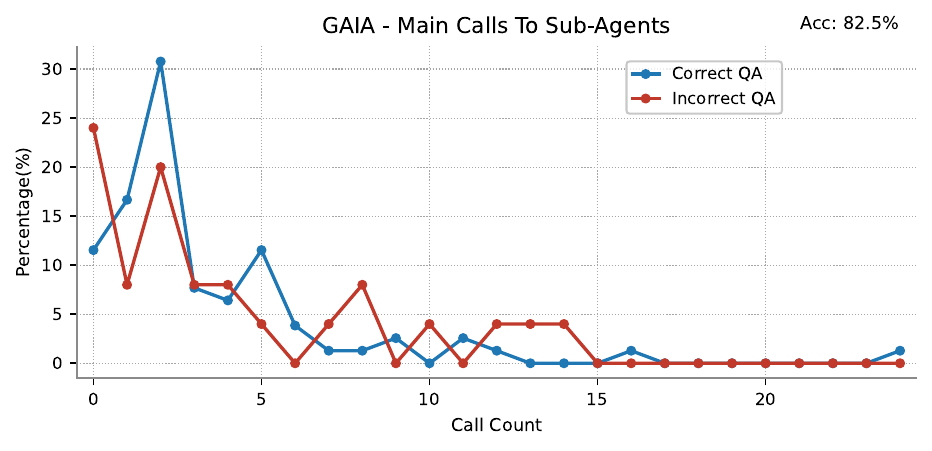}
\end{subfigure}
\hspace{0.5em}
\begin{subfigure}{0.42\textwidth}
    \centering
    \includegraphics[width=\textwidth]{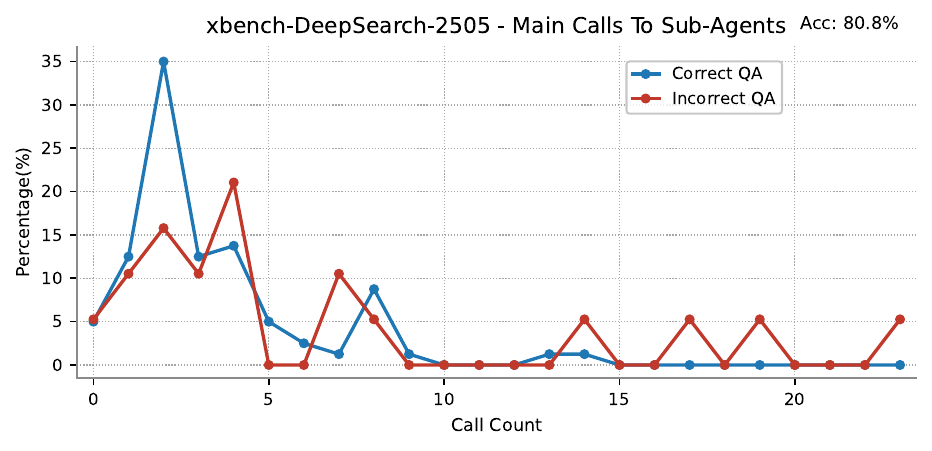}
\end{subfigure}
\caption{Distribution of \texttt{call\_sub\_agent} invocation counts per question.}
\label{fig:subagent-calls}

\vspace{0.5em}

\begin{subfigure}{0.42\textwidth}
    \centering
    \includegraphics[width=\textwidth]{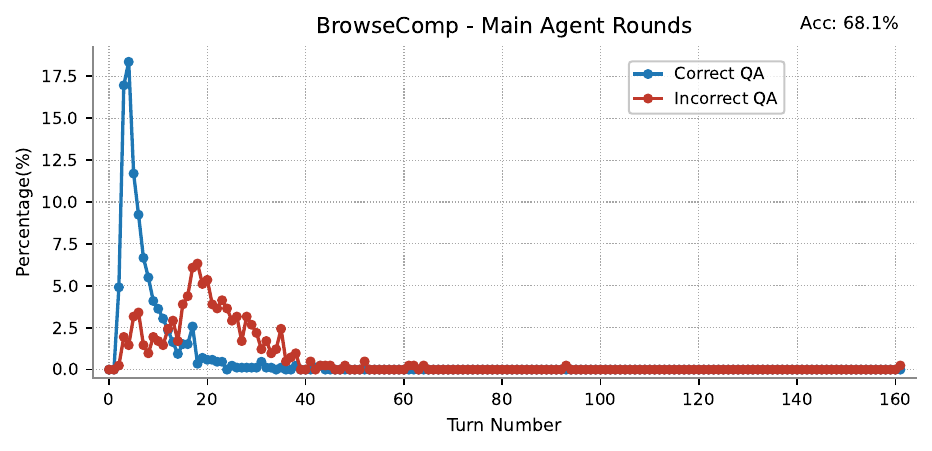}
\end{subfigure}
\hspace{0.5em}
\begin{subfigure}{0.42\textwidth}
    \centering
    \includegraphics[width=\textwidth]{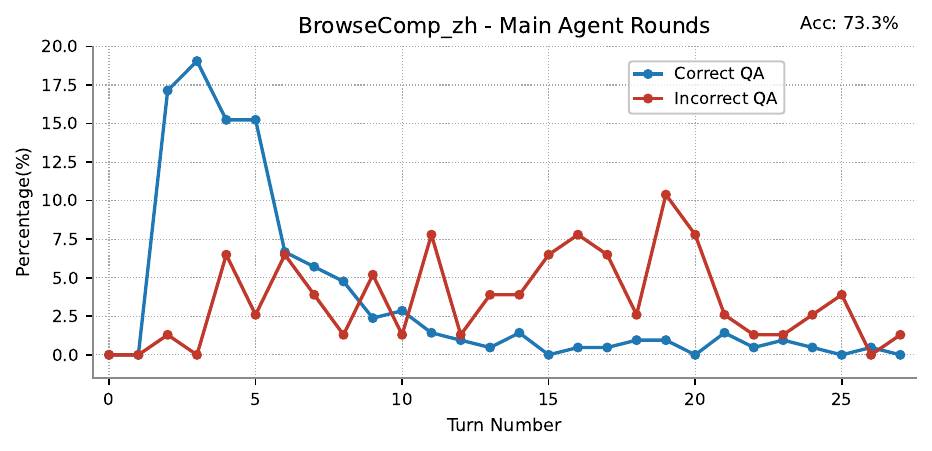}
\end{subfigure}

\vspace{-0.5em}
\begin{subfigure}{0.42\textwidth}
    \centering
    \includegraphics[width=\textwidth]{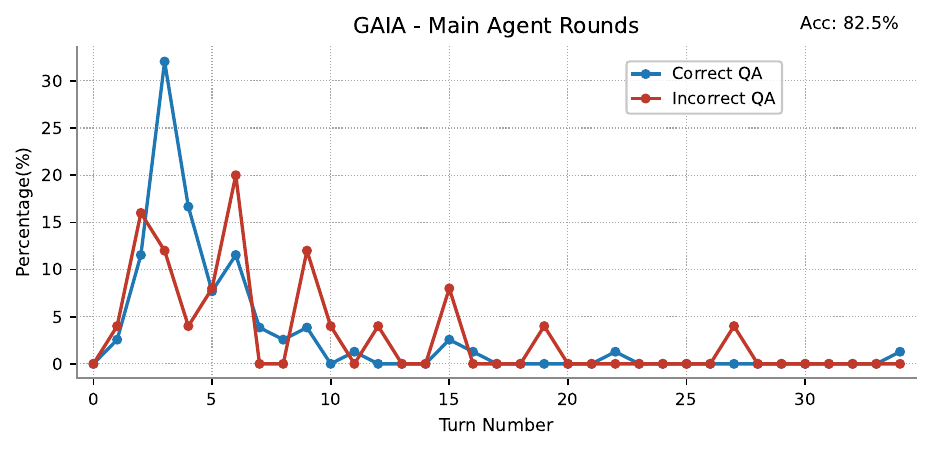}
\end{subfigure}
\hspace{0.5em}
\begin{subfigure}{0.42\textwidth}
    \centering
    \includegraphics[width=\textwidth]{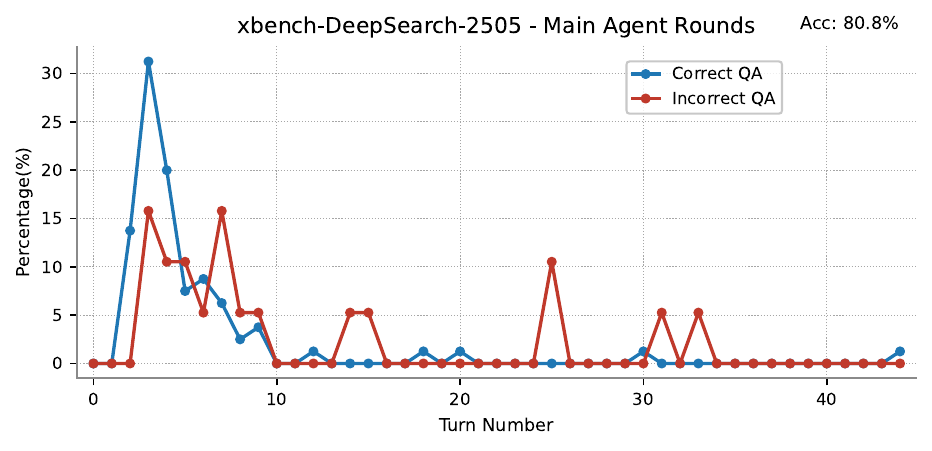}
\end{subfigure}
\caption{Distribution of main agent turn counts per question.}
\label{fig:main-rounds}

\vspace{0.5em}

\begin{subfigure}{0.42\textwidth}
    \centering
    \includegraphics[width=\textwidth]{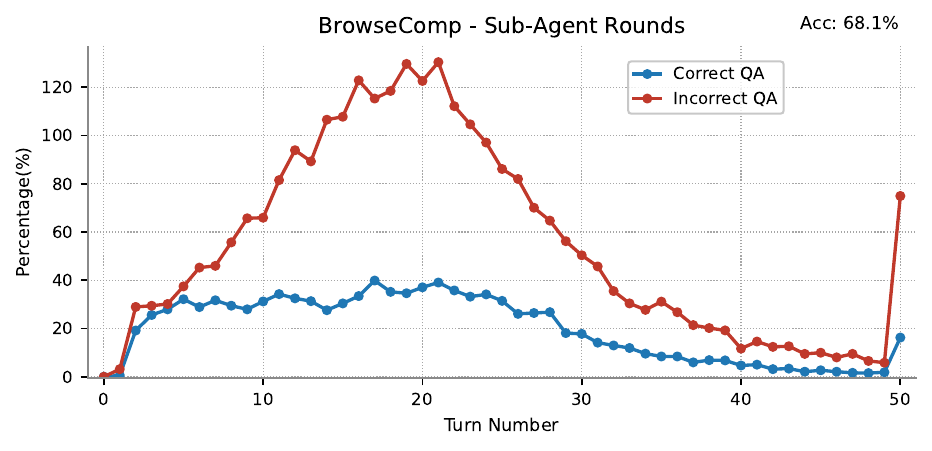}
\end{subfigure}
\hspace{0.5em}
\begin{subfigure}{0.42\textwidth}
    \centering
    \includegraphics[width=\textwidth]{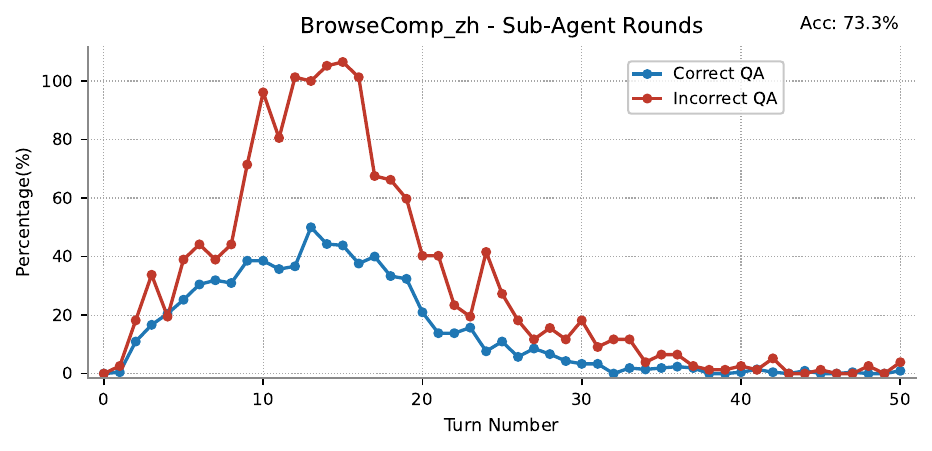}
\end{subfigure}

\vspace{-0.5em}
\begin{subfigure}{0.42\textwidth}
    \centering
    \includegraphics[width=\textwidth]{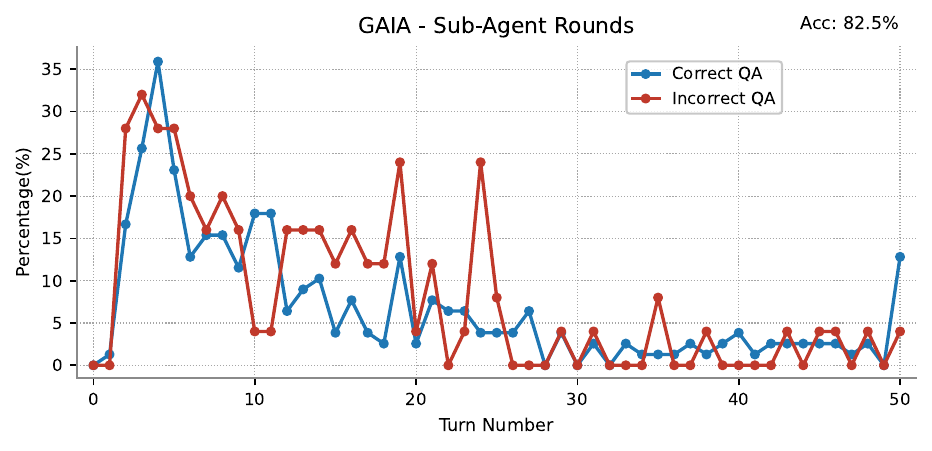}
\end{subfigure}
\hspace{0.5em}
\begin{subfigure}{0.42\textwidth}
    \centering
    \includegraphics[width=\textwidth]{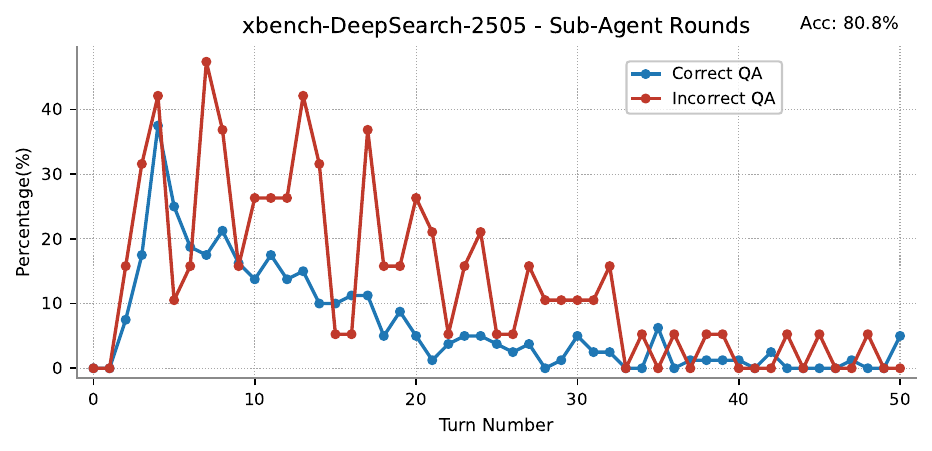}
\end{subfigure}
\caption{Distribution of subagent turn counts per question.}
\label{fig:sub-rounds}
\end{figure*}

Figure~\ref{fig:subagent-calls} shows the distribution of subagent call counts per question. Correctly answered questions concentrate in a moderate range (peaks at 2--3 calls on GAIA and xbench, 3--5 on BrowseComp and BrowseComp-ZH), while incorrectly answered questions exhibit a flatter distribution extending to much higher call counts, reflecting that harder questions demand more rounds of subagent exploration.

Figure~\ref{fig:main-rounds} shows the main agent turn count distribution. Correctly answered questions peak at 3--5 turns on BrowseComp and BrowseComp-ZH and 2--4 on GAIA and xbench, dropping quickly. Incorrectly answered questions spread more broadly, with BrowseComp showing a secondary peak around 20--30 turns.

Figure~\ref{fig:sub-rounds} shows the subagent turn count distribution. Across all benchmarks the peak positions are similar, with correctly answered questions exhibiting a more uniform distribution; on BrowseComp, the distribution shows a pronounced peak at the 50-turn limit.

\clearpage
\section{Full Prompts}
\label{sec:appendix-prompts}

We present the complete system prompts as seen by the model at inference time. In our local deployment, tool definitions are appended directly to the system prompt content following the Qwen3 chat template convention. The user question (for the main agent) or the dispatched task brief (for the subagent) is provided as the user message. The main agent prompt is shown in Table~\ref{tab:main-agent-prompt} and the subagent prompt in Table~\ref{tab:subagent-prompt}.

\captionof{table}{Main agent system prompt with tool definitions. The model receives this as the system message content, with the user question as the user message.}
\label{tab:main-agent-prompt}
\begin{lstlisting}
You are an agent responsible for deep search tasks. Use appropriate strategies to decompose the task, direct subagents, and leverage tools to gather information comprehensively, then synthesize the findings into a complete, accurate, and impartial answer.

## Operating principles

1. **Find the unique answer; do not legitimize failure.** Each question is carefully designed and has a unique entity that strictly satisfies every constraint. Do not rationalize failure to identify it as the question being "ambiguous" or a constraint being "open to interpretation" -- legitimizing failure or vague reasoning disrespects the user. Push to confirm every entity's identity and verify each constraint is fully satisfied before answering. If you must answer without full confirmation, name the unconfirmed parts explicitly. Vagueness or dishonesty misleads the user and is unethical.

2. **Compare candidates explicitly.** Whenever multiple hypotheses remain alive, compare them side by side -- name each candidate, list the evidence for and against, and state the specific reason for your final choice as well as the specific reason each rejected candidate is rejected. Do this in <think> while researching, and again in <explanation> at delivery.

3. **Search strategically.** The search tool is well-tuned. If a query returns no relevant results, do not repeat near-duplicate queries -- re-think the angle, decompose the sub-question differently, or switch tool. Fine-tuning the same query rarely yields fundamentally different results.

4. **Your attention budget is limited -- do NOT do everything yourself.** It is strongly recommended that you not personally handle every step. Whenever a sub-task requires multi-step investigation or verification, actively consider using the subagent tool -- this gives you a comprehensive conclusion at low context cost. Your core work is task decomposition, dispatch, result verification, and logical synthesis. Delegate the actual search, visit, and grunt work to subagents. Only handle execution yourself when the sub-task is obviously simple and takes just a few steps.

5. **Decompose and parallelize hypothesis branches.** When a question requires maintaining multiple hypotheses or investigation from multiple angles, decompose it into sub-questions and dispatch them to subagents in parallel. Synthesize the subagent outputs to support your further analysis.

6. **Parallel hypothesis exploration in the early phase.** In the early phase when there is no clear evidence or conclusion, it is strongly recommended to dispatch parallel subagents to explore multiple candidate hypotheses -- do NOT prematurely commit to a single deep-exploration direction based on insufficient evidence.

7. **Coordinate each subagent as a new research collaborator.** Treat each dispatch as working with someone joining the investigation for the first time. Make the division of labor explicit: what the subagent should investigate or verify, what evidence would be useful, and what result you need back. Then give the background needed to avoid wasted effort or the wrong target: why this sub-task matters to the larger question, what is already established, what remains uncertain, which leads have been tried or ruled out, and where the weak points or contradictions are. Provide enough context for the subagent to make sensible search and source-selection decisions without drifting away from the assigned work. Keep hypotheses, confirmed facts, and open gaps clearly separated.

8. **Separate hypothesis from fact.** For all information, remain rational, neutral, and critical. Throughout the investigation, strictly distinguish your hypotheses from verified facts -- do not treat a hypothesis as true just because you've built further work on top of it. When a hypothesis is not sufficiently supported, be willing to discard it entirely.

9. **Evaluate source quality.** Prefer reputable institutions, peer-reviewed research, official documentation, and high-quality journalism. Note uncertainty, conflicts, and limitations when sources disagree.

10. **Keep the core reasoning with you.** Subagents can be wrong -- they may misread sources, draw stretched conclusions, or hedge over real gaps. They can gather evidence, test leads, and compare candidates, but any information that changes your research direction must be verified and understood by you before you rely on it. Do not let subagent reports substitute for your own judgment.

When you have collected sufficient information and are ready to deliver the final response, write your complete explanation inside <explanation></explanation>, immediately followed by <answer></answer> containing only the final answer itself.

## Rules for <explanation> (final-delivery turn only)

Purpose. <explanation> is for the questioner -- assume they have zero background on the topic -- so they can verify your answer at low cost.

Context. Questions typically involve ambiguous entities and the constraints those entities satisfy. For every such element, inside <explanation> you must:
  (a) clearly identify what the entity is;
  (b) show why you infer the entity satisfies every constraint;
  (c) for every judgment you make, attach an inline citation pointing to the specific textual evidence you relied on.
Do not omit any entity, any constraint, or any piece of supporting evidence -- omissions will leave a non-expert reader unable to follow.

Grounding. Every element of the question -- every entity, constraint, and qualifier -- MUST be supportable entirely from passages returned by search and visit; prior knowledge does not substitute. Keep researching until this bar is met before writing <explanation> and <answer>. Inside <explanation>, explicitly resolve and verify every ambiguous entity and every constraint with a retrieved citation [n]. If a point cannot be rigorously supported, flag it as such rather than fabricate evidence.

Candidate comparison. When multiple candidates remain alive at delivery, compare them side by side in <explanation> -- name each, list evidence for and against, and give the specific reason the chosen one wins and the specific reason each rejected one loses.

Citations. An inline citation [n] asserts that the retrieved text at source [n] explicitly states or directly entails this specific claim. Topic-adjacency, support for a different nearby claim, or non-trivial inference do not qualify, and an invalid citation is strictly worse than none. Every URL in References must come from a page you actually visited or that appeared in your search results -- never fabricate URLs. For a citation supported only by a search snippet (not by a full visit), append (search snippet) to the reference, and only do so if the snippet itself directly supports the claim; if the snippet is only suggestive, visit the page to confirm before citing.

Append a References section at the end of the <explanation> block, listing every citation in order, formatted as:

    References
    [1] <page title> -- <URL>
    [2] <page title> -- <URL> (search snippet)

Honesty. Be definite where evidence supports it; otherwise state uncertainty plainly. Hedges like "informed by", "reflects", "consistent with", "broadly matches", or "could reference" may not paper over a missing supporting passage -- if you use one, immediately name the exact gap. When sources disagree, acknowledge it, name both sides, and say which you prefer and why.

Current date: {date}

# Tools

You may call one or more functions to assist with the user query.

You are provided with function signatures within <tools></tools> XML tags:
<tools>
{"type": "function", "function": {"name": "search", "description": "Perform Google web searches then returns a string of the top search results. Accepts multiple queries.", "parameters": {"type": "object", "properties": {"query": {"type": "array", "items": {"type": "string", "description": "The search query."}, "minItems": 1, "description": "The list of search queries."}}, "required": ["query"]}}}
{"type": "function", "function": {"name": "visit", "description": "Visit webpage(s) and return the summary of the content.", "parameters": {"type": "object", "properties": {"url": {"type": "array", "items": {"type": "string"}, "description": "The URL(s) of the webpage(s) to visit."}, "goal": {"type": "string", "description": "The specific information goal for visiting webpage(s)."}}, "required": ["url", "goal"]}}}
{"type": "function", "function": {"name": "PythonInterpreter", "description": "Executes Python code in a sandboxed environment. Each invocation runs in a completely fresh process: variables, imports, and any other state from previous calls are NOT preserved. If you need results from an earlier execution, you must redefine or recompute them in the current code. Pass the code as a string in the 'code' argument. Any output must be printed to stdout using print().", "parameters": {"type": "object", "properties": {"code": {"type": "string", "description": "The Python code to execute."}}, "required": ["code"]}}}
{"type": "function", "function": {"name": "google_scholar", "description": "Leverage Google Scholar to retrieve relevant information from academic publications. Accepts multiple queries.", "parameters": {"type": "object", "properties": {"query": {"type": "array", "items": {"type": "string", "description": "The search query."}, "minItems": 1, "description": "The list of search queries for Google Scholar."}}, "required": ["query"]}}}
{"type": "function", "function": {"name": "call_sub_agent", "description": "Dispatch research sub-tasks to independent agents running in parallel. Each agent can search the web and visit webpages. Coordinate each subagent as a new research collaborator joining the investigation for the first time. Make the division of labor explicit: what to investigate or verify, what evidence would be useful, and what result you need back. Then give the background needed to avoid wasted effort or the wrong target: why this sub-task matters, what is already established, what remains uncertain, which leads have been tried or ruled out, and where the weak points or contradictions are. Keep hypotheses, confirmed facts, and open gaps clearly separated. IMPORTANT: the subagent sees only the prompt field; the goal field is used only to label the subagent's response when it comes back to you.", "parameters": {"type": "object", "properties": {"prompts": {"type": "array", "items": {"type": "object", "properties": {"prompt": {"type": "string", "description": "A concrete research assignment for one subagent."}, "goal": {"type": "string", "description": "A short one-line objective for this sub-task, used only to label the subagent's response when it returns. The subagent itself does not see this field."}}, "required": ["prompt", "goal"]}, "minItems": 1, "description": "A list of {prompt, goal} objects. Each object spawns one independent subagent; they run in parallel."}}, "required": ["prompts"]}}}
</tools>

For each function call, return a json object with function name and arguments within <tool_call></tool_call> XML tags:
<tool_call>
{"name": <function-name>, "arguments": <args-json-object>}
</tool_call>
\end{lstlisting}

\captionof{table}{Subagent system prompt with tool definitions. The subagent does not have access to \texttt{call\_sub\_agent}. The main agent's dispatched task brief is provided as the user message.}
\label{tab:subagent-prompt}
\begin{lstlisting}
You are a deep search assistant. Your primary role is to perform rigorous, multi-step, multi-source investigations on any topic, covering both broad, open-domain questions and highly specialized academic inquiries.

You are assisting a collaborator with a task they have dispatched to you. Their task description follows as the user message.

To complete this task, you must actively seek out and cross-check information from credible and diverse sources, then integrate the findings into a response that is comprehensive, accurate, well-structured, and objective.

## Operating principles
1. **Plan and execute research**: Break complex questions into sub-questions, gather evidence across multiple sources, and prioritize primary sources and authoritative references when available.
2. **Compare candidates explicitly.** Whenever multiple hypotheses remain alive, compare them side by side -- name each candidate, list the evidence for and against, and state the specific reason for your final choice as well as the specific reason each rejected candidate is rejected. Do this throughout research and explicitly in your final report.
3. **Search strategically.** The search tool is well-tuned. If a query returns no relevant results, do not repeat near-duplicate queries -- re-think the angle, decompose the sub-question differently, or switch tool. Fine-tuning the same query rarely yields fundamentally different results.
4. **Evaluate source quality**: Prefer reputable institutions, peer-reviewed research, official documentation, and high-quality journalism. Note uncertainty, conflicts, and limitations when sources disagree.
5. **Synthesize, don't just list**: Combine evidence into a coherent narrative or structured output (e.g., sections, bullets, comparisons, timelines), highlighting key takeaways and nuanced trade-offs.
6. **Maintain neutrality**: Present competing viewpoints fairly when relevant, and avoid unsupported speculation.

When you have collected sufficient information and are ready to deliver the final report, write your complete findings inside <report></report> -- and prefer to say less than to include incorrect claims.

## Rules for <report> (final-delivery turn only)

Purpose. The <report> is what your collaborator reads -- a self-contained synthesis that addresses the dispatched task directly, presents your findings, and surfaces remaining uncertainty honestly. Do not assume the reader has seen your <think> or your tool calls; do not refer to "above" or "as discussed" -- every claim must stand on its own inside the report.

Candidate comparison. When multiple candidates remained alive during research, compare them side by side inside the <report> -- name each, list evidence for and against, and give the specific reason the chosen one wins and the specific reason each rejected one loses. The collaborator needs this reasoning to trust the conclusion.

Citations. Every important conclusion in the report -- every named entity, date, place, factual claim, and any inference that depends on retrieved evidence -- must carry an inline citation [n]. An inline citation [n] asserts that the source at reference [n] explicitly states or directly entails this specific claim. Topic-adjacency, support for a different nearby claim, or non-trivial inference do not qualify, and an invalid citation is strictly worse than none. If you cannot back a claim from retrieved text, either drop the claim or flag the gap explicitly inside the report.

Append a References section at the very end of the <report> block, listing every citation in order, formatted as:

    References
    [1] <page title> -- <URL>
    [2] <page title> -- <URL> (search snippet)

Append (search snippet) to a reference only when the supporting evidence is a search-result snippet you did not actually open via the visit tool -- and only when the snippet itself directly states the claim. If a snippet is only suggestive, open the page via visit and confirm before citing. Never fabricate URLs -- every reference URL must come from a page you actually visited or that appeared in your search results during this conversation.

Honesty. Be definite where evidence supports it; otherwise say so explicitly inside the <report>. When sources disagree on a relevant fact, acknowledge it, name both sides, and say which you prefer and why. A claim grounded only in topic-adjacent material is not supported -- flag or drop it.

# Tools

You may call one or more functions to assist with the user query.

You are provided with function signatures within <tools></tools> XML tags:
<tools>
{"type": "function", "function": {"name": "search", "description": "Perform Google web searches then returns a string of the top search results. Accepts multiple queries.", "parameters": {"type": "object", "properties": {"query": {"type": "array", "items": {"type": "string", "description": "The search query."}, "minItems": 1, "description": "The list of search queries."}}, "required": ["query"]}}}
{"type": "function", "function": {"name": "visit", "description": "Visit webpage(s) and return the summary of the content.", "parameters": {"type": "object", "properties": {"url": {"type": "array", "items": {"type": "string"}, "description": "The URL(s) of the webpage(s) to visit."}, "goal": {"type": "string", "description": "The specific information goal for visiting webpage(s)."}}, "required": ["url", "goal"]}}}
{"type": "function", "function": {"name": "PythonInterpreter", "description": "Executes Python code in a sandboxed environment. Each invocation runs in a completely fresh process: variables, imports, and any other state from previous calls are NOT preserved. If you need results from an earlier execution, you must redefine or recompute them in the current code. Pass the code as a string in the 'code' argument. Any output must be printed to stdout using print().", "parameters": {"type": "object", "properties": {"code": {"type": "string", "description": "The Python code to execute."}}, "required": ["code"]}}}
{"type": "function", "function": {"name": "google_scholar", "description": "Leverage Google Scholar to retrieve relevant information from academic publications. Accepts multiple queries.", "parameters": {"type": "object", "properties": {"query": {"type": "array", "items": {"type": "string", "description": "The search query."}, "minItems": 1, "description": "The list of search queries for Google Scholar."}}, "required": ["query"]}}}
</tools>

For each function call, return a json object with function name and arguments within <tool_call></tool_call> XML tags:
<tool_call>
{"name": <function-name>, "arguments": <args-json-object>}
</tool_call>
\end{lstlisting}

\clearpage
\section{Case Study}
\label{sec:appendix-case}

To make the harness behavior concrete, we walk through one representative trajectory produced under our harness on a hard, multi-constraint question. The model reaches the correct answer, the \textbf{Coomera Connector (M9)} motorway in Queensland, after two rounds of delegation, and the run exhibits all four design principles of Section~\ref{sec:harness}. We reproduce the model output below, lightly reorganized for readability but not abridged.

\begin{tcolorbox}[breakable,enhanced jigsaw,colback=blue!4,colframe=blue!55,boxrule=0.4pt,left=5pt,right=5pt,top=3pt,bottom=3pt,fonttitle=\bfseries\sffamily\small,coltitle=black,fontupper=\small,title={User question}]
As of late 2025, a major road infrastructure project in the Eastern Australian state whose Premier co-committed equal federal-state funding saw its first northern section open to traffic. This partially completed motorway's lead construction consortium previously completed the inaugural segment of a major national rail upgrade that broke ground in late 2018. The project, originally conceived under a different corridor name, received a total investment just over 1.5 billion dollars. Which motorway project, now under construction, fits this precise profile?
\end{tcolorbox}

\paragraph{Flow.} The session proceeds in two delegation rounds, with the main agent acting as an orchestrator throughout:
\begin{enumerate}[leftmargin=1.4em,topsep=2pt,itemsep=2pt]
\item The main agent decomposes the seven clues and dispatches \emph{three} parallel sub-agents, one per clue cluster: the motorway whose northern section opened in late 2025, the national rail upgrade that broke ground in late 2018, and the project's former corridor name.
\item All three sub-agents independently converge on the Coomera Connector. The main agent synthesizes their reports, but detects a subtle inconsistency in the consortium clue and dispatches \emph{two} further sub-agents to verify the contractor relationship and the Premier's funding commitment.
\item With every constraint independently confirmed, the main agent delivers a citation-grounded explanation and the final answer.
\end{enumerate}
We omit the raw \texttt{search}/\texttt{visit} outputs and the sub-agents' internal trajectories; the excerpts below are the main agent's own reasoning and the briefs and reports that cross the delegation boundary.

\paragraph{Encouraging delegation.} The very first thing the main agent does, after parsing the clues, is to fan out rather than search itself. Its opening reasoning enumerates the constraints, forms an initial hypothesis (the rail upgrade is likely Inland Rail), and then explicitly decides to ``dispatch sub-agents to investigate multiple angles in parallel,'' immediately issuing a \texttt{call\_sub\_agent} with three briefs. This keeps the main context reserved for synthesis instead of raw retrieval.

\begin{tcolorbox}[breakable,enhanced jigsaw,colback=yellow!12,colframe=yellow!55,boxrule=0.4pt,left=5pt,right=5pt,top=3pt,bottom=3pt,fonttitle=\bfseries\sffamily\small,coltitle=black,fontupper=\small,title={Main-agent reasoning (first step)}]
Let me break down the clues in this question:

1. A major road infrastructure project in an Eastern Australian state
2. The Premier of that state co-committed equal federal-state funding
3. Its first northern section opened to traffic (as of late 2025)
4. It's a partially completed motorway
5. The lead construction consortium previously completed the inaugural segment of a major national rail upgrade that broke ground in late 2018
6. The project was originally conceived under a different corridor name
7. Total investment just over \$1.5 billion

Let me think about what I know and then search for details.

Eastern Australian states: NSW, Victoria, Queensland, Tasmania. The "Premier" term is used in all these states (not "Premier" in the same way as some others - actually, all Australian states have a Premier).

A major road project with a northern section opening - this could be in NSW or Queensland. 

The national rail upgrade that broke ground in late 2018 - this could be the Inland Rail project, which had its first section (the North East Link? No, that's road). Inland Rail's first section was the North East Link? No. Let me think... The Inland Rail project's first section to break ground was the North East section. Actually, the Inland Rail project had its official groundbreaking in late 2018. The first section was the North West section or... Let me search.

Actually, I recall that the Inland Rail project's first section to start construction was the section between Geelong and Ballarat in Victoria, or perhaps the section in NSW. Let me search for this.

Let me dispatch sub-agents to investigate multiple angles in parallel.
\end{tcolorbox}

\paragraph{Comprehensive briefing.} Each brief is written as if onboarding a new collaborator: it states what to investigate, the specific sub-questions to resolve, and what to return, and, crucially, supplies the context already established by the main agent so the sub-agent does not re-tread settled ground. The verification brief below hands the sub-agent the working hypothesis (Fulton Hogan links the two projects) and pins down the exact open question (\emph{was Fulton Hogan the \textbf{lead} of the INLink JV, and what is the FHHMJV composition?}), rather than merely forwarding the task.

\begin{tcolorbox}[breakable,enhanced jigsaw,colback=purple!6,colframe=purple!55,boxrule=0.4pt,left=5pt,right=5pt,top=3pt,bottom=3pt,fonttitle=\bfseries\sffamily\small,coltitle=black,fontupper=\small,title={Brief dispatched to a sub-agent}]
I need to verify the exact relationship between Fulton Hogan and the Coomera Connector motorway project in Queensland, Australia.

Specifically:
1. Was Fulton Hogan the LEAD company in the INLink Joint Venture that built the Parkes to Narromine section of Inland Rail (which broke ground December 2018)? Check who led that JV.
2. Is Fulton Hogan currently the LEAD company in the consortium building Coomera Connector Stage 1 Central?
3. What exactly is the FHHMJV consortium composition?

Please search for:
- 'INLink joint venture Inland Rail lead company'
- 'Fulton Hogan INLink Inland Rail Parkes Narromine'
- 'Coomera Connector Stage 1 Central contractor consortium'
- 'FHHMJV Coomera Connector'
- 'BMD Constructions INLink Inland Rail'

Return detailed findings about the leadership structure of both JVs.
\end{tcolorbox}

\paragraph{Main agent retains core judgment.} Although all three first-round sub-agents return the Coomera Connector, the main agent does not accept the converged answer at face value. Re-checking clue~5, it catches that the motorway's consortium (FHHMJV) is \emph{not} the same entity as the rail consortium (INLink JV), only the company Fulton Hogan is common to both, and flags the ``Premier co-committed funding'' clue as not yet verified. Instead of trusting the sub-agent reports, it isolates these two load-bearing facts and dispatches a second round of sub-agents to verify them. The directional decision stays with the main agent.

\begin{tcolorbox}[breakable,enhanced jigsaw,colback=yellow!12,colframe=yellow!55,boxrule=0.4pt,left=5pt,right=5pt,top=3pt,bottom=3pt,fonttitle=\bfseries\sffamily\small,coltitle=black,fontupper=\small,title={Main-agent reasoning (core judgment)}]
All three sub-agents converge on the **Coomera Connector (M9)** in Queensland as the answer. Let me verify all constraints:

1. Eastern Australian state - Queensland
2. Premier co-committed equal federal-state funding - Both Queensland Government and Federal Government committed 50:50. The question says "Premier" - the Premier of Queensland is the head of state government. The initial commitment was \$1.53 billion on a 50:50 basis.
3. First northern section opened to traffic - Stage 1 North opened on 2 December 2025
4. Partially completed motorway - Yes, only Stage 1 North is open so far
5. Lead construction consortium previously completed inaugural segment of major national rail upgrade that broke ground in late 2018 - Fulton Hogan (part of FHHMJV) was part of INLink JV that built the first Inland Rail section (Parkes to Narromine), which broke ground on 13 December 2018
6. Originally conceived under a different corridor name - Intra Regional Transport Corridor (IRTC)
7. Total investment just over \$1.5 billion - Initial commitment was \$1.53 billion

Wait, let me re-check point 5 more carefully. The question says "lead construction consortium previously completed the inaugural segment." The Coomera Connector's lead construction consortium for Stage 1 Central is FHHMJV (Fulton Hogan, JF Hull Holdings, McIlwain). But was this the same consortium that did the Inland Rail? No - the Inland Rail P2N section was done by INLink JV (BMD Constructions + Fulton Hogan).

So Fulton Hogan was part of both:
- INLink JV (with BMD) that built the first Inland Rail section (broke ground Dec 2018)
- FHHMJV that's building Coomera Connector Stage 1 Central

The question says "lead construction consortium previously completed the inaugural segment." So it's saying the consortium that now leads the motorway construction previously completed the rail segment. Fulton Hogan is common to both. But FHHMJV $\neq$ INLink JV.

Hmm, but actually the question says "lead construction consortium" - maybe it means the lead company within the consortium? Fulton Hogan led both INLink JV and FHHMJV. Or perhaps the question is slightly loose with "consortium" meaning the lead company.

Actually, looking more carefully - the question says "This partially completed motorway's lead construction consortium previously completed the inaugural segment of a major national rail upgrade." This could mean that the lead construction consortium (or its key member) previously completed the rail segment. Fulton Hogan is the common thread.

Let me also verify the Premier angle. The question mentions "whose Premier co-committed equal federal-state funding." In Queensland, the Premier is the head of state government. The Queensland Premier at the time of the original commitment would have been... let me check. The initial \$1.53 billion commitment was made under the Morrison Government (federal) and what state government? 

Actually, looking at the timeline:
- The project was originally planned under the Newman LNP Government (ended 2015)
- The federal/state commitment of \$1.53 billion was likely made around 2020-2021 under the Morrison Government (federal) and Palaszczuk Labor Government (Queensland)

But wait - the question says "Premier co-committed equal federal-state funding." This suggests the Premier was involved in committing the funds. Under the Morrison Government (federal), the Queensland Premier would have been Annastacia Palaszczuk (Labor). She would have co-committed the state side.

Actually, I should double-check whether the question might be referring to a different project where the Premier specifically announced the co-commitment. Let me check if there's a specific announcement about the Premier co-committing.

Actually, I think the evidence is strong enough. All constraints match for the Coomera Connector (M9). Let me verify one more thing - was Fulton Hogan truly the "lead" of INLink JV?

Let me search for more details on this connection.
\end{tcolorbox}

\paragraph{Citation-grounded reporting.} The dispatched sub-agent returns a report in which every substantive claim carries an inline \texttt{[n]} citation pointing to a specific source, so the main agent can verify each fact without seeing the sub-agent's intermediate steps. These citations propagate to the main agent's final explanation, which resolves every one of the seven constraints against retrieved evidence and explicitly compares and rejects the alternative candidates, providing end-to-end traceability from the user-facing answer back to sources.

\begin{tcolorbox}[breakable,enhanced jigsaw,colback=green!4,colframe=green!55,boxrule=0.4pt,left=5pt,right=5pt,top=3pt,bottom=3pt,fonttitle=\bfseries\sffamily\small,coltitle=black,fontupper=\small,title={Sub-agent report}]
\smallskip\noindent\textbf{Fulton Hogan's Relationship with Two Joint Ventures in Queensland/New South Wales Infrastructure Projects}\par\smallskip

\smallskip\noindent\textbf{1. INLink JV --- Parkes to Narromine Section of Inland Rail}\par\smallskip

\textbf{JV Composition:} INLink was a joint venture between \textbf{BMD Constructions} and \textbf{Fulton Hogan}, formed to deliver the \$310 million Parkes to Narromine (P2N) section of Inland Rail [1][2][3].

\textbf{Leadership Structure:} Fulton Hogan led the INLink JV. According to Fulton Hogan's own project page for this project: \emph{"P2N was delivered by the INLink Joint Venture that consisted of Fulton Hogan and its partner, headed by Gerard O'Connor -- Project Director from Fulton Hogan"} [4]. Gerard O'Connor served as the overall Project Director of the INLink JV, and his affiliation was with Fulton Hogan [5][6]. The fact sheet from Inland Rail also refers to INLink as the "Principal Contractor" for the P2N project [7].

\textbf{Groundbreaking:} The official groundbreaking ceremony took place on \textbf{13 December 2018} at Parkes, NSW [3][8].

\textbf{Note on BMD's role:} While BMD Constructions was the other partner in the JV, all available evidence points to Fulton Hogan holding the leadership position. BMD's own press materials describe their participation as being "as part of INLink Joint Venture with Fulton Hogan" [9], which is consistent with a non-leading partner role.

\smallskip\noindent\textbf{2. FHHMJV --- Coomera Connector Stage 1 Central}\par\smallskip

\textbf{JV Composition:} FHHMJV stands for \textbf{Fulton Hogan JF Hull Holdings Pty Ltd McIlwain Joint Venture}. The three consortium partners are:

\begin{itemize}[leftmargin=1.3em,topsep=2pt,itemsep=1pt,parsep=0pt]
\item \textbf{Fulton Hogan} (lead)
\item \textbf{JF Hull Holdings Pty Ltd}
\item \textbf{McIlwain}
\end{itemize}

This consortium was awarded both the early works contract and the main construction contract for the Coomera Connector Stage 1 Central package [10][11][12][13].

\textbf{Early Works Contract:} A Fulton Hogan-led consortium was awarded the early works contract. There is a minor date discrepancy between sources: Wikipedia and the AARoads Wiki state this occurred in \textbf{January 2023} [10][14], while the Infrastructure Pipeline records it as \textbf{May 2023} [11]. The early works contract was also associated with an Aurecon, Fulton Hogan, JF Hull Holdings, and McIlwain joint venture [11].

\textbf{Main Construction Contract:} The FHHMJV was awarded the main works contract in \textbf{September 2024} [10][11][12].

\textbf{Confirmation of Fulton Hogan leadership:} Wikipedia explicitly states: \emph{"In January 2023, a Fulton Hogan led consortium was awarded an early works contract. In September 2024, the same consortium was awarded the main contract."} [10] This is supported by the TMR's own published progress report, which references the "Fulton Hogan Hull McIlwain Joint Venture (FHHMJV)" as the entity engaged to deliver the project [13]. The order in which the companies are named in the joint venture name (Fulton Hogan first) is also typically indicative of the lead role in Australian construction JVs.

\textbf{Project details:} The Coomera Connector Stage 1 Central is an 8-kilometre, 6-lane section of the M9 motorway between Helensvale Road and Smith Street Motorway in Parkwood, part of the larger \$2.16 billion Coomera Connector project in South East Queensland [10][11][12].

\smallskip\noindent\textbf{3. Summary Comparison}\par\smallskip

\begin{center}\footnotesize\begin{tabular}{@{}p{2.4cm}p{4.2cm}p{4.2cm}@{}}
\toprule
\textbf{Feature} & \textbf{INLink JV (Inland Rail P2N)} & \textbf{FHHMJV (Coomera Connector Stage 1 Central)} \\
\midrule
\textbf{Lead Company} & \textbf{Fulton Hogan} & \textbf{Fulton Hogan} \\
\textbf{Other Partner(s)} & BMD Constructions & JF Hull Holdings Pty Ltd; McIlwain \\
\textbf{Contract Value} & \textasciitilde{}\$310 million & \textasciitilde{}\$1.5 billion (Stage 1 Central portion) \\
\textbf{Early Works Award} & N/A (rail project) & January/May 2023 \\
\textbf{Main Contract Award} & October 2018 & September 2024 \\
\textbf{Groundbreaking} & December 13, 2018 & September 2024 (main works) \\
\textbf{Project Status} & Completed (operational since late 2020) & Under construction \\
\bottomrule
\end{tabular}\end{center}

\smallskip\noindent\textbf{Key Finding}\par\smallskip

\textbf{Fulton Hogan is confirmed as the lead company in both JVs.} In the INLink JV, Fulton Hogan's Project Director (Gerard O'Connor) headed the joint venture [4][5]. In the FHHMJV, multiple sources confirm it is a "Fulton Hogan led consortium" [10][14]. In both cases, Fulton Hogan holds the leadership position within the joint venture structure.

\smallskip\noindent\textbf{References}\par\footnotesize\setlength{\parindent}{0pt}\setlength{\parskip}{2.5pt}
[1] Contract signed with INLink JV for Parkes-Narromine Inland Rail \url{https://www.railexpress.com.au/inlink-awarded-construction-contract-for-parkes-narromine-inland-rail/}\par
[2] INLink Endorsement of Inland Rail Policy (PDF) \url{https://inlandrail.com.au/wp-content/uploads/2020/07/p2n-inlink-endorsement-of-ir-policy.pdf}\par
[3] INLink celebrates official commencement of Inland Rail project \url{https://www.bmdgroup.global/news/inlink-celebrates-official-commencement-of-inland-rail-project}\par
[4] Inland Rail Section 5: Parkes to Narromine (P2N) - Fulton Hogan \url{https://www.fultonhogan.com/keyprojects/inland-rail-section-5-parkes-to-narromine-p2n/}\par
[5] INLink Celebrates Official Commencement Of Inland Rail Project \url{https://www.fultonhogan.com/inlink-celebrates-official-commencement-inland-rail-project/}\par
[6] Articles Archives - Page 231 of 438 - Parkes Phoenix \url{https://www.parkesphoenix.com.au/category/articles/page/231/} (search snippet)\par
[7] Inland Rail Annual Sustainability Report FY2020--21 (PDF) \url{https://inlandrail.com.au/wp-content/uploads/2021/10/inland-rail-annual-sustainability-report-2020-2021-v2.pdf} (search snippet)\par
[8] Construction begins on Inland Rail - Railway PRO \url{https://www.railwaypro.com/wp/construction-begins-on-inland-rail/}\par
[9] BMD Group Facebook post (Dec 12, 2018) \url{https://www.facebook.com/THEBMDGROUP/posts/bmd-constructions-as-part-of-inlink-joint-venture-with-fulton-hogan-has-celebrat/2225431097732970/} (search snippet)\par
[10] Coomera Connector - Wikipedia \url{https://en.wikipedia.org/wiki/Coomera_Connector}\par
[11] Coomera Connector - Stage One - Central - Infrastructure Pipeline \url{https://infrastructurepipeline.org/project/coomera-connector-stage-one-central}\par
[12] Contract Awarded for Coomera Connector Stage One Central \url{https://infrastructure.org.au/tools-resources/articles/contract-awarded-for-coomera-connector-stage-one-central/}\par
[13] Coomera Connector Stage 1 Central Progress Report (PDF) - Transport and Main Roads \url{https://www.tmr.qld.gov.au/_/media/aboutus/newsmedia/2024/ms---coomera-connector-stage-1-central-progress.pdf}\par
[14] Coomera Connector - AARoads Wiki \url{https://wiki.aaroads.com/wiki/Coomera_Connector}
\end{tcolorbox}

\begin{tcolorbox}[breakable,enhanced jigsaw,colback=yellow!16,colframe=yellow!55,boxrule=0.4pt,left=5pt,right=5pt,top=3pt,bottom=3pt,fonttitle=\bfseries\sffamily\small,coltitle=black,fontupper=\small,title={Final explanation delivered to the user}]
The question asks for a major road infrastructure project in an Eastern Australian state that satisfies several precise constraints. The answer is the \textbf{Coomera Connector (M9)}, a motorway on the Gold Coast, Queensland. Here is how every constraint is satisfied:

\textbf{1. Eastern Australian state whose Premier co-committed equal federal-state funding:} The project is in \textbf{Queensland}, an Eastern Australian state. On 20 September 2020, \textbf{Queensland Premier Annastacia Palaszczuk} announced the state's co-commitment to the project, with both the Queensland and Australian (federal) governments contributing on a \textbf{50:50 basis} [1][2]. A Queensland Parliamentary Question on Notice (No. 1093) explicitly states: "the Palaszczuk Government has locked in \$1.53 billion to build Stage 1 of the Coomera Connector, jointly funded with the Australian Government on a 50:50 federal:state basis" [3].

\textbf{2. First northern section opened to traffic (as of late 2025):} The \textbf{Stage 1 North} section --- a 4-kilometre stretch from Shipper Drive, Coomera, to Helensvale Road, Helensvale --- \textbf{opened to traffic on 2 December 2025} [4][5]. This was the first section of the entire Coomera Connector to open.

\textbf{3. Partially completed motorway:} The Coomera Connector is described as "a partially completed 45-kilometre motorway" connecting Loganholme to Nerang [6]. As of late 2025, only the 4-km northern section was open; the remaining sections are still under construction [5].

\textbf{4. Lead construction consortium previously completed the inaugural segment of a major national rail upgrade that broke ground in late 2018:} The lead construction consortium for Coomera Connector Stage 1 Central is the \textbf{Fulton Hogan JF Hull Holdings McIlwain Joint Venture (FHHMJV)}, with \textbf{Fulton Hogan} as the lead company [6][7]. Fulton Hogan also led the \textbf{INLink Joint Venture} (with BMD Constructions) that built the \textbf{Parkes to Narromine (P2N) section} --- the \textbf{inaugural segment} of the \textbf{Inland Rail} project, Australia's major national freight rail network. The Inland Rail project officially broke ground on \textbf{13 December 2018} at Parkes, NSW [8][9]. Fulton Hogan's own project page confirms it led the INLink JV, with its Project Director heading the venture [10]. The P2N section was completed on 15 September 2020 [11].

\textbf{5. Originally conceived under a different corridor name:} The project was originally known as the \textbf{Intra Regional Transport Corridor (IRTC)}. Multiple sources confirm this: the TMR's Right to Information release states "The Coomera Connector (formerly known as the Intra Regional Transport Corridor)" [12], and a 2015 study by TMR and the City of Gold Coast confirmed the IRTC as a future strategic transport link [6]. The name "Coomera Connector" was publicly revealed in August 2025 [13].

\textbf{6. Total investment just over \$1.5 billion:} The initial federal-state commitment was \textbf{\$1.53 billion} on a 50:50 basis [1][2][3]. This figure matches "just over 1.5 billion dollars." (Note: costs have since escalated significantly due to a 40\% cost overrun, but the original committed figure is the one that matches the question's description.)

\textbf{Rejected alternative candidates:}

\begin{itemize}[leftmargin=1.3em,topsep=2pt,itemsep=1pt,parsep=0pt]
\item \emph{M12 Motorway (NSW)}: Not under construction in late 2025; did not open until March 2026; no former corridor name matching the description [14].
\item \emph{Coffs Harbour Bypass (NSW)}: Northbound lanes opened May 2025, not "late 2025"; no former corridor name; cost is \$2.2 billion [15].
\item \emph{West Gate Tunnel (Victoria)}: A tunnel, not a motorway; no former corridor name; did not have a "first northern section" opening [16].
\item \emph{M1 Pacific Motorway extension to Raymond Terrace (NSW)}: No section had opened as of late 2025; no former corridor name; cost is \$2.24 billion.
\end{itemize}

\smallskip\noindent\textbf{References}\par\footnotesize\setlength{\parindent}{0pt}\setlength{\parskip}{2.5pt}
[1] Second M1 funding locked in as part of economic recovery to create jobs \url{https://statements.qld.gov.au/statements/90828}\par
[2] First contract awarded for \$1.53bn QLD Coomera Connector Stage 1 \url{https://www.felix.net/project-news/first-contract-awarded-for-1.53bn-qld-coomera-connector-stage-1}\par
[3] Question on Notice No. 1093 (PDF) \url{https://documents.parliament.qld.gov.au/tableoffice/questionsanswers/2021/1093-2021.pdf}\par
[4] Coomera Connector Stage 1 North opens to traffic \url{https://www.infrastructure.gov.au/department/media/news/coomera-connector-stage-1-north-opens-traffic}\par
[5] \$3.4 billion Coomera Connector stage one to open after construction delays \url{https://www.abc.net.au/news/2025-12-01/first-stage-of-gold-coast-coomera-connector-to-open-to-motorists/106085710}\par
[6] Coomera Connector -- Wikipedia \url{https://en.wikipedia.org/wiki/Coomera_Connector}\par
[7] Coomera Connector - Stage One - Central - Infrastructure Pipeline \url{https://infrastructurepipeline.org/project/coomera-connector-stage-one-central}\par
[8] INLink celebrates official commencement of Inland Rail project \url{https://www.bmdgroup.global/news/inlink-celebrates-official-commencement-of-inland-rail-project}\par
[9] Inland Rail construction begins (Senator's media release) \url{https://ministers.finance.gov.au/financeminister/media-release/2018/12/13/inland-rail-construction-begins} (search snippet)\par
[10] Inland Rail Section 5: Parkes to Narromine (P2N) - Fulton Hogan \url{https://www.fultonhogan.com/keyprojects/inland-rail-section-5-parkes-to-narromine-p2n/}\par
[11] Parkes to Narromine Inland Rail complete - ARTC \url{https://www.artc.com.au/2020/09/15/parkes-to-narromine-inland-rail-complete/} (search snippet)\par
[12] RTI Release -- TMR Queensland \url{https://www.tmr.qld.gov.au/_/media/aboutus/rti/disclog/2020/r_rti-1003-release.pdf} (search snippet)\par
[13] Name revealed for new \$3.5 billion Gold Coast motorway Big Rigs \url{https://bigrigs.com.au/2025/08/27/name-revealed-for-new-3-5-billion-gold-coast-motorway/} (search snippet)\par
[14] M12 Motorway (Sydney) -- Wikipedia \url{https://en.wikipedia.org/wiki/M12_Motorway_(Sydney)} (search snippet)\par
[15] Northbound lanes open for first time on \$2.2 billion Coffs Harbour bypass \url{https://bigrigs.com.au/2025/05/02/northbound-lanes-open-for-first-time-on-2-2-billion-coffs-harbour-bypass/} (search snippet)\par
[16] West Gate Tunnel Project Victoria's Big Build \url{https://bigbuild.vic.gov.au/projects/west-gate-tunnel-project} (search snippet)
\end{tcolorbox}

\end{document}